\documentclass[10pt,twocolumn,letterpaper]{article}

\usepackage[pagenumbers]{cvpr} 

\usepackage{multirow}
\usepackage{makecell}
\usepackage{adjustbox} 

\definecolor{cvprblue}{rgb}{0.21,0.49,0.74}
\usepackage[pagebackref,breaklinks,colorlinks,allcolors=cvprblue]{hyperref}

\def\paperID{10480} 
\def\confName{CVPR}
\def\confYear{2026}

\title{Cross-Modal Emotion Transfer for Emotion Editing in Talking Face Video}

\author{
Chanhyuk Choi \qquad
Taesoo Kim \qquad
Donggyu Lee \qquad
Siyeol Jung \qquad
Taehwan Kim \\
Ulsan National Institute of Science and Technology (UNIST) \\
Ulsan, Republic of Korea \\
{\tt\small \{chan4184, taesoo0630, leedongkyu2019, siyeol, taehwankim\}@unist.ac.kr}
}

\begin{document}
\maketitle
\begin{abstract}
Talking face generation has gained significant attention as a core application of generative models.
To enhance the expressiveness and realism of synthesized videos, emotion editing in talking face video plays a crucial role.
However, existing approaches often limit expressive flexibility and struggle to generate extended emotions.
Label-based methods represent emotions with discrete categories, which fail to capture a wide range of emotions.
Audio-based methods can leverage emotionally rich speech signals—and even benefit from expressive text-to-speech (TTS) synthesis—but they fail to express the target emotions because emotions and linguistic contents are entangled in emotional speeches.
Images-based methods, on the other hand, rely on target reference images to guide emotion transfer, yet they require high-quality frontal views and face challenges in acquiring reference data for extended emotions (e.g., sarcasm).
To address these limitations, we propose \textbf{Cross-Modal Emotion Transfer (C-MET)}, a novel approach that generates facial expressions based on speeches by modeling emotion semantic vectors between speech and visual feature spaces.
C-MET leverages a large-scale pretrained audio encoder and a disentangled facial expression encoder to learn emotion semantic vectors that represent the difference between two different emotional embeddings across modalities.
Extensive experiments on the MEAD and CREMA-D datasets demonstrate that our method improves emotion accuracy by 14\% over state-of-the-art methods, while generating expressive talking face videos—even for unseen extended emotions. 
Code, checkpoint, and demo are available at \url{https://chanhyeok-choi.github.io/C-MET/}.
\end{abstract}

\vspace{-2.0em}    
\section{Introduction}
\label{sec:intro}

Talking face generation has recently achieved significant advancements, with numerous methods developed to synthesize realistic videos driven by audio inputs~\cite{zhou2020makelttalk,chen2019hierarchical,das2020speech,zakharov2019few,zhong2023identity,wang2021audio2head,wang2022one,chen2020talking,yang2022face2face}. This research area enables a wide range of applications, including virtual human animation, filmmaking, and digital entertainment~\cite{pataranutaporn2021ai}. Early studies primarily focused on improving visual fidelity, preserving speaker identity, and ensuring accurate lip synchronization~\cite{das2020speech,zhong2023identity,wang2021audio2head}. More recently, the field has shifted toward emotional talking face generation~\cite{ji2022eamm,gan2023efficient,tan2025edtalk,ki2024float}, aiming to enhance the expressiveness and naturalness of generated videos but often focusing on generating basic emotions~\cite{ekman1992argument}. Synthesizing complex and subtle emotional expressions is essential for creating believable virtual agents~\cite{liu2025moee, chen-etal-2024-emoknob}, as it significantly improves human–computer interaction, fosters emotional engagement, and enables more immersive and empathetic communication in applications such as education, therapy, and virtual assistants~\cite{hernandez2023affective,rings2024empathy,saffaryazdi2025empathetic}.

\begin{figure}[t]
  \centering
  \includegraphics[width=\linewidth]{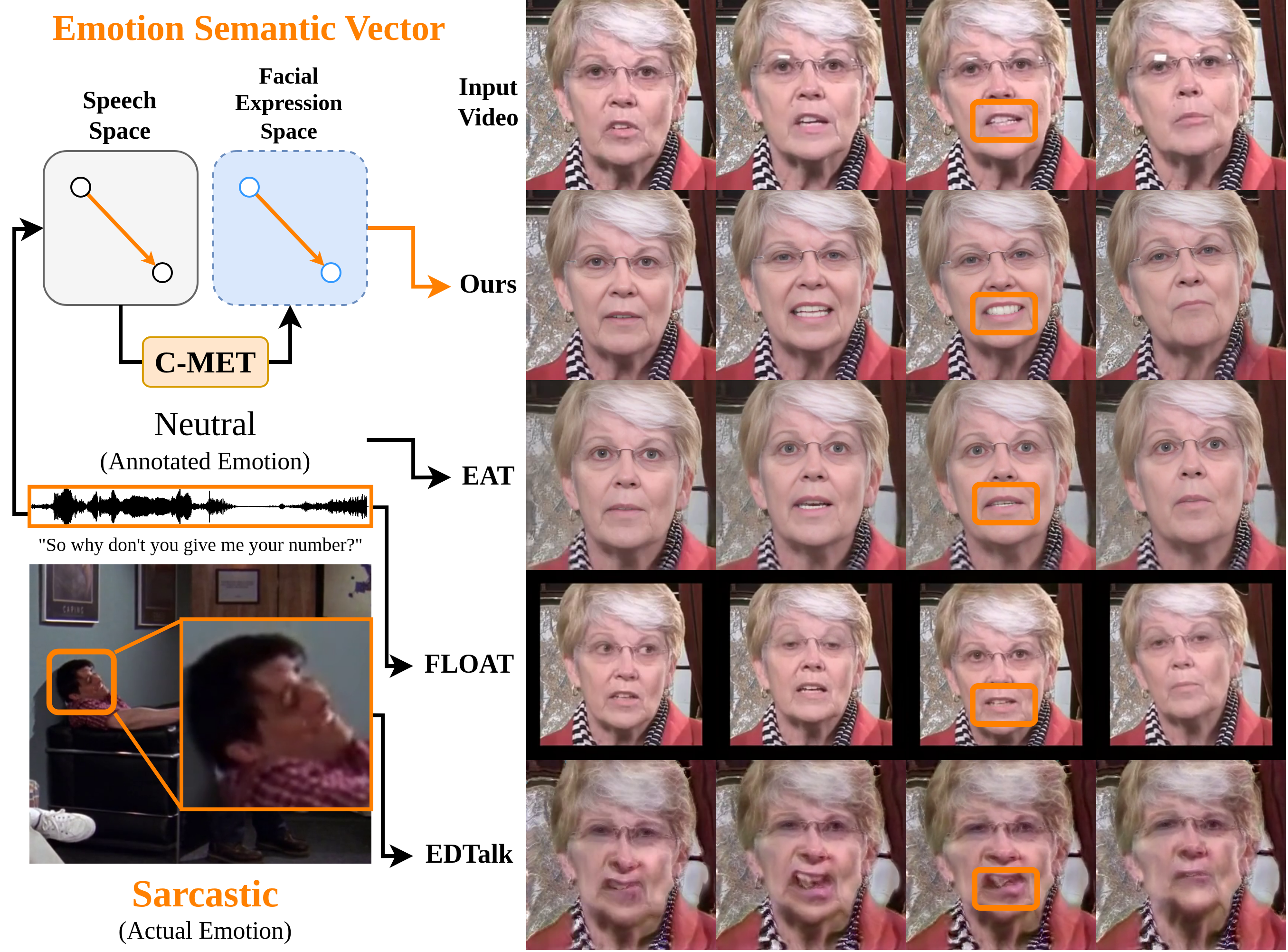}
  \vspace{-1.75em} 
  \caption{\textbf{Comparison between our method and baseline approaches.}
Identity, lip, and pose are taken from a neutral video, while the emotion source is provided from MELD~\cite{poria2018meld} (dialogue 5, utterance 8).
From top to bottom: ours (C-MET), the label-based method (EAT~\cite{gan2023efficient}), the existing audio-based method (FLOAT~\cite{ki2024float}), and the image-based method (EDTalk~\cite{tan2025edtalk}).
Our results better reflect the target emotional speech (\textit{sarcastic}), exhibiting a more pronounced widening of the lip corners compared to the baselines.}
  \label{fig:teaser}
  \vspace{-1.6em} 
\end{figure}

\begin{figure}[t]
  \centering
  \includegraphics[width=\linewidth]{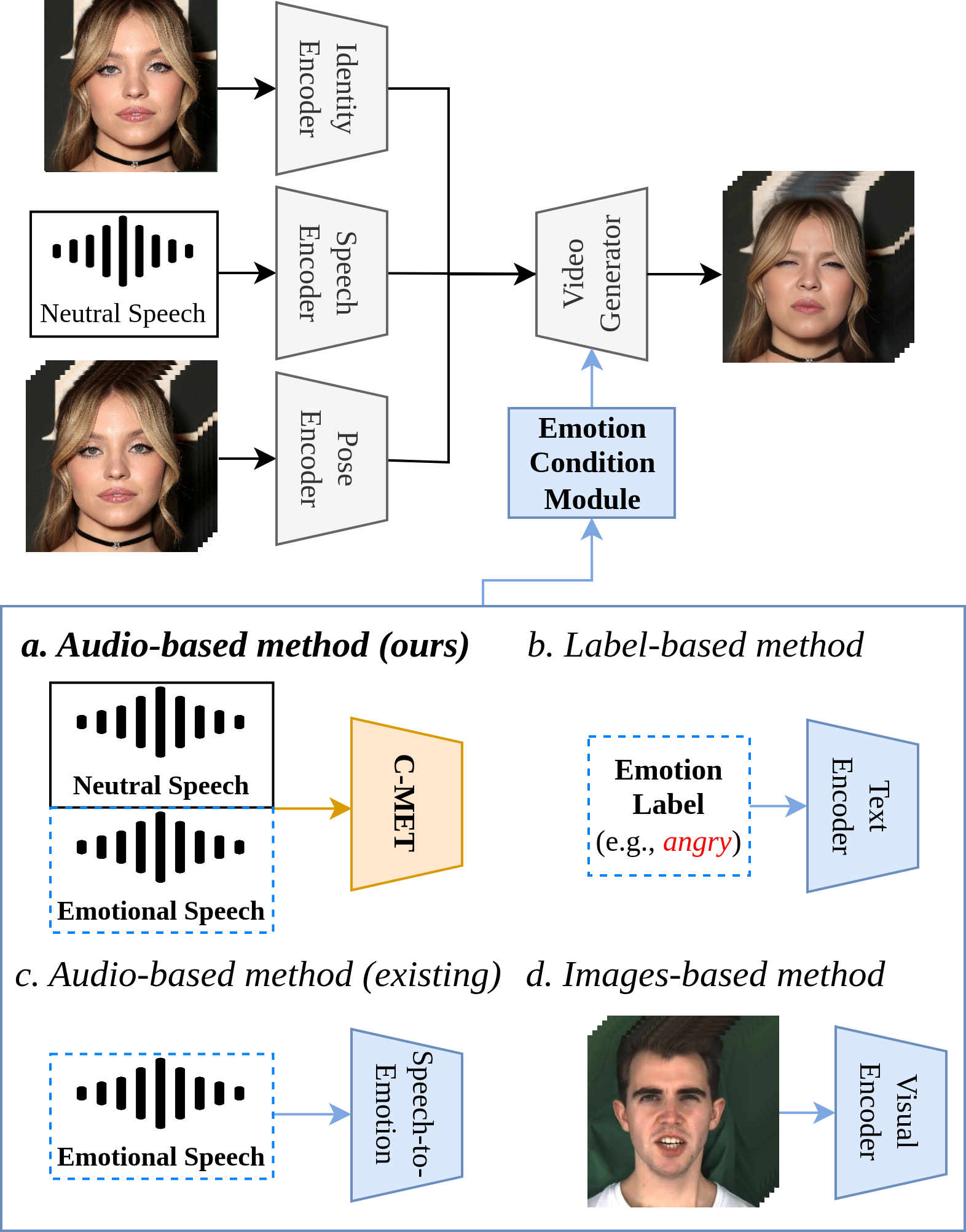}
  \vspace{-1.75em} 
  \caption{The comparison of emotion condition modules in the pipeline of emotion editing in talking face video at inference.}
  \vspace{-1.6em} 
  \label{fig:emotion_editing}
\end{figure}

One recent line of research in emotional talking face generation~\cite{sun2023continuously} decomposes this task into two sub-tasks: generating an emotionless talking face video and editing the facial expressions in each frame using external emotion signals.
As illustrated in Figure~\ref{fig:emotion_editing}, the second sub-task, emotion editing in talking face video, typically uses one of three modalities as emotion sources: (1) discrete emotion labels, (2) emotional speeches, or (3) reference images. Label-based methods~\cite{goyal2023emotionally,gan2023efficient,sun2023continuously,wang2024eat,lin2025mvportrait,liu2025moee} rely on a limited set of categorical labels (e.g., eight basic emotions), restricting scalability and expressiveness.
Audio-based methods~\cite{tan2023emmn,shen2024emotalker,ki2024float,dai2025emohuman,shen2025emohead,tian2024emo,liu2025moee} typically struggle to disentangle emotion from linguistic content, as illustrated in Figure~\ref{fig:teaser}.
Images-based methods~\cite{ji2022eamm,tan2025edtalk,wang2025emotivetalk,liu2025moee} can achieve stronger emotional fidelity by directly referencing expressive facial images, but they require frontal-view reference samples and substantial preprocessing, which limits usability. Moreover, generating extended emotions~\cite{chen-etal-2024-emoknob} (e.g., \textit{sarcasm}, \textit{charisma}) requires large-scale audio-visual paired data with the emotion labels~\cite{liu2025moee}.
Expressing extended emotions in talking-face video without collecting additional audio-visual paired datasets thus remains a non-trivial challenge. On the other hand, thanks to the recent progress in expressive text-to-speech (TTS) systems~\cite{comanici2025gemini}, we can easily synthesize complex emotional speech. Furthermore, emotional speech databases are also readily accessible~\cite{busso2008iemocap, lotfian2017building}.
Therefore, motivated by the advance in fine-grained voice cloning~\cite{chen-etal-2024-emoknob}, we leverage such emotional audios to extract emotion semantic vectors that capture rich affective cues—independent of lip-motion signals, unlike prior audio-based methods.
However, bridging the domain gap between audio and visual emotion representations remains difficult, making cross-modal mapping a key open problem.

To address the aforementioned challenges, we propose a novel \textbf{Cross-Modal Emotion Transfer (C-MET)} learning method, to the best of our knowledge, which is the first to explicitly model the relationship of \emph{emotion semantic vectors} between audio and visual feature spaces.
Here, an emotion semantic vector is obtained by subtracting the embedding of two different emotional expressions.
C-MET learns to predict the semantic vectors in the visual space from those in the audio space, effectively transferring emotion semantics across modalities.
Specifically, given input audios, our method first extracts emotion semantic vectors from the audio space using a self-supervised emotion representation model pretrained on large-scale speech data~\cite{ma2023emotion2vec}, and then predicts corresponding emotion semantic vectors in the visual space via a facial expression encoder from a state-of-the-art disentanglement framework~\cite{tan2025edtalk}.
Inspired by extended voice emotion control from Emoknob~\cite{chen-etal-2024-emoknob}, we extend this idea to the visual domain and generation task by mapping emotion semantic vectors in speech and facial expression spaces.
Leveraging the rich and continuous emotion representations inherent in audio, our method enables the synthesis of extended emotions that were never observed during training.
Moreover, our method can be seamlessly integrated as a plug-and-play module into existing disentanglement-based talking face generators, enhancing emotional expressiveness while reducing inference time by replacing the heavy facial expression encoder with our lightweight module.
 
We evaluate our method on the MEAD~\cite{wang2020mead} and CREMA-D~\cite{cao2014crema} datasets, covering a wide range of emotions and speaking styles.
Experimental results demonstrate that our method effectively learns emotion-specific transformations in cross-modal space and validate its ability to generate extended emotions. Our method achieves notable improvements over state-of-the-art methods, as evidenced by both quantitative and qualitative evaluations.
The contributions of our work can be summarized as follows:
\begin{itemize}
    \item We propose C-MET, to the best of our knowledge, the first approach for emotion editing in talking face video that enables extended emotional talking face generation by modeling the mapping between emotion semantic vectors of large-scale speech feature space and facial expression feature space.
    \item We propose a novel and simple yet effective cross-modal transformer module to generate emotion semantic vectors in facial expression space, which can be used as a plug-and-play module into existing disentanglement-based talking face generation models.
    \item Our method narrows the modality gap between speech and facial expressions by cross-modal emotion transfer and can generate extended emotional talking face video even for unseen emotions, as validated by both quantitative metrics and qualitative assessments.
\end{itemize}
\section{Related Work}

\subsection{Audio-driven Talking Face Generation}
Audio-driven talking face generation has become a prominent topic 
in generative modeling, with extensive research focused on achieving 
realism and precise audio–lip synchronization while preserving the 
identity of the reference image~\cite{chen2019hierarchical,zakharov2019few,das2020speech,zhou2020makelttalk,liu2022semantic,bregler2023video, chen2025echomimic, xu2024vasa}.
Early works such as 
Wav2Lip~\cite{prajwal2020lip} blend synthesized lip movements into 
existing frames, though they occasionally exhibit visual artifacts 
around the mouth area. Subsequent two-stage approaches~\cite{zhong2023identity,wang2021audio2head,wang2022one,chen2020talking,yang2022face2face} predict intermediate representations from audio before reconstructing 
facial frames, but tend to capture only coarse motion patterns and 
introduce accumulated errors that degrade visual realism. 
In contrast, reconstruction-based approaches~\cite{chen2018lip,chen2023vast,shen2022learning,shen2023difftalk,thies2020neural,wang2023lipformer,zhou2019talking} integrate 
multimodal features in an end-to-end manner, alleviating the 
aforementioned issues. In particular, disentanglement 
frameworks~\cite{tan2025edtalk,zhou2021pose,pang2023dpe,yu2023talking,
wang2023progressive} represent accurate facial dynamics—such as head 
pose, lip motion, and emotional expressions—in a global 
manner~\cite{tan2023emmn}. Although these models can capture facial 
expressions across different identities via disentangled expression 
encoders, their performance remains optimal only when the identity 
of the driving and target faces is consistent. Since the distribution of facial expressions varies across identities, 
our model learns average emotional representations by sampling within 
the same speaker identity during training, thereby isolating emotion 
from identity-specific facial variations in the disentangled space. Accordingly, our approach aims to generate 
expressive facial dynamics directly from emotional audio and 
integrate them into disentanglement-based networks to enhance 
emotional talking face generation.

\subsection{Emotion Editing in Talking Face Video}

More recently, emotional talking face generation has gained attention for producing realistic and expressive talking face video~\cite{yin2022styleheat, liang2022expressive, sun2024fg, peng2023emotalk, ji2022eamm,gan2023efficient,tan2025edtalk,ki2024float}. To do this, after synthesizing a talking face video for precise lip synchronization, properly synthesizing a target emotion into each frame is introduced by Sun et al.~\cite{sun2023continuously}. 
To control facial expressions, existing methods typically condition the generation process on specific control signals such as emotion labels, emotional audio, or driving images.

EAT~\cite{gan2023efficient} introduces a lightweight transformer-based adaptation network that controls emotions using discrete emotion labels, while Style2Talker~\cite{tan2024style2talker} employs a diffusion model combined with CLIP~\cite{radford2021learning} to inject textual emotion descriptions into a 3DMM-based generation pipeline.
However, label-based methods~\cite{goyal2023emotionally,gan2023efficient,sun2023continuously,wang2024eat,lin2025mvportrait} are inherently limited to predefined emotion categories, making it difficult to represent extended or subtle emotional states. Audio-based methods~\cite{tan2023emmn,shen2024emotalker,ki2024float,dai2025emohuman,shen2025emohead,tian2024emo} use a speech as both the lip-motion and emotion source. For instance, FLOAT~\cite{ki2024float} employs a speech-to-emotion module to redirect the target emotion, but fails to accurately reflect the intended emotion when the lip-sync audio and emotional source differ—indicating that speech content and emotional cues are not fully disentangled. Images-based methods~\cite{ji2022eamm,tan2025edtalk,wang2025emotivetalk,liu2025moee} attempt to overcome this limitation by directly referencing expressive images. EAMM~\cite{ji2022eamm} generates keypoint motions from reference images but suffers from identity inconsistency and limited expressive control. StyleTalk~\cite{ma2023styletalk} develops a style encoder to capture the style of a reference video. EDTalk~\cite{tan2025edtalk} improves upon this by introducing a disentanglement framework that separates lip motion, head pose, and facial expressions using emotional video sources as driving signals.
While this enables more flexible expression control, it still relies heavily on curated, high-quality emotional video inputs.
Moreover, such methods often fail to capture nuanced or underrepresented emotions (e.g., \textit{sarcastic}, \textit{charismatic}) due to the lack of diverse emotional video data. Recently, MoEE~\cite{liu2025moee} attempts to address complex emotion generation through a mixture-of-emotion-experts framework, yet it still requires a large amount of additional labeled data for predefined complex emotions.

In contrast, our method tackles these limitations using only the MEAD dataset~\cite{wang2020mead} by leveraging a large-scale representation audio encoder and a facial expression encoder of disentanlgement networks. Therefore, we introduce an intermediate network that learns to generate emotion semantic vectors in the facial expression space conditioned on emotion semantic vectors in the audio space. This design not only reduces the modality gap for accurate cross-modal emotion regression but also enables the generation of unseen emotional expressions.

\begin{figure*}[t]
  \centering
  \includegraphics[width=0.95\textwidth]{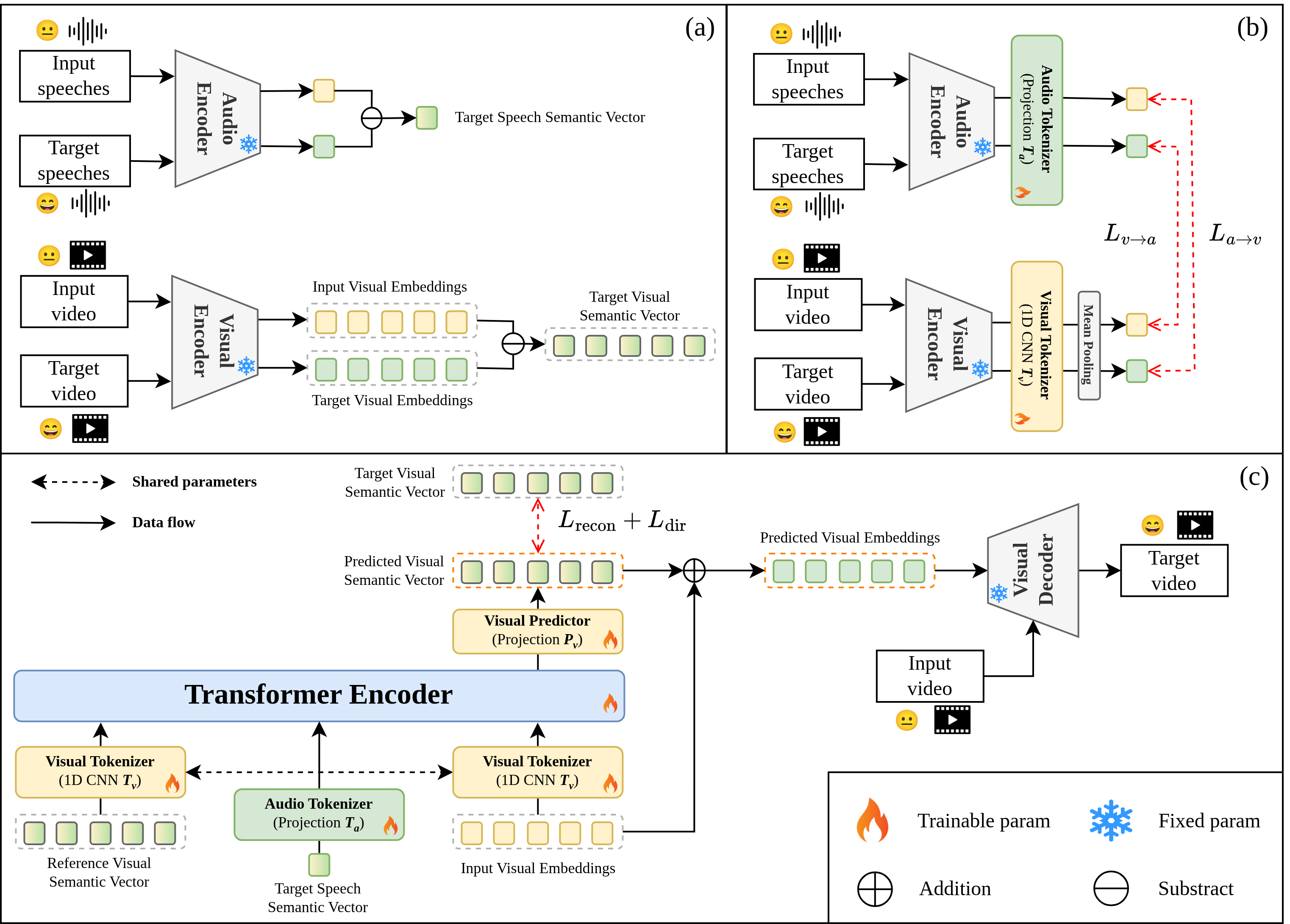}
  \vspace{-0.5em} 
  \caption{\textbf{Overview of the proposed Cross-Modal Emotion Transfer (C-MET).} (a) We extract input and target embeddings using pretrained audio and visual encoders, and compute the semantic vectors by subtracting the target embeddings from the inputs. (b) During training, we apply contrastive learning between multimodal tokens—both from visual to audio and audio to visual—to align the representation spaces. (c) A multimodal transformer encoder is used to regress the target expression vectors, guided by the speech vectors. The predicted vectors are then added to the input visual embeddings, which are decoded by a pretrained visual decoder to reconstruct the target video from the neutral video.}
  \label{fig:overall architecture}
  \vspace{-1.5em} 
\end{figure*}
\section{Methodology}

Given input and target pairs of audios \(A\) and videos \(V\), our objective is to learn the correlation between emotion semantic vectors across seperate audio and visual spaces. Once trained, the model predicts target visual semantic vectors from input visual embeddings, guided by the corresponding semantic vectors in the audio domain. These semantic vectors are subsequently used to synthesize emotional talking face videos.
As illustrated in Figure~\ref{fig:overall architecture}, we extract 
modality-specific embeddings using pretrained encoders, align them 
via learnable tokenizers in a shared latent space, and pass them to 
a Transformer encoder to model cross-modal correspondence.
Specifically, emotion2vec+large~\cite{ma2023emotion2vec} is adopted as the pretrained audio encoder, while the facial expression encoder of EDTalk~\cite{tan2025edtalk} is used as the pretrained visual encoder.
To disentangle multimodal tokens between emotions, we employ a contrastive learning objective during training.
As shown in Figure~\ref{fig:overall architecture}-(c), the Transformer encoder predicts the target visual semantic vectors in the facial expression space.
Finally, the predicted vectors are added to the input visual embeddings to obtain the target visual embedding, which are fed into the pretrained visual decoder to synthesize the emotional talking face video. 

\begin{table*}[t]
\centering
\setlength{\tabcolsep}{1.1mm}
\renewcommand{\arraystretch}{1.05}
\adjustbox{max width=\textwidth}{
\begin{tabular}{l|c|ccccc|ccccc}
\toprule
 & \textbf{Emotion} & \multicolumn{5}{c|}{\textbf{MEAD}} & \multicolumn{5}{c}{\textbf{CREMA-D}} \\ 
\makecell[c]{\textbf{Method}} & \textbf{Source Type} &
AITV\,$\downarrow$ & FID\,$\downarrow$ & FVD\,$\downarrow$ &
$\text{Sync}_{\text{conf}}\,\uparrow$ & $\text{Acc}_{\text{emo}}\,\uparrow$ &
AITV\,$\downarrow$ & FID\,$\downarrow$ & FVD\,$\downarrow$ &
$\text{Sync}_{\text{conf}}\,\uparrow$ & $\text{Acc}_{\text{emo}}\,\uparrow$ \\ 
\midrule
EAMM~\cite{ji2022eamm} & Images & 3.745 & 161.602 & 474.446 & 6.0609 & 18.81 & 6.481 & 206.168 & 628.344 & 4.1134 & 19.15 \\
EAT~\cite{gan2023efficient} & Label  & 12.575 & 90.974 & 330.722 & \underline{8.0528} & 41.56 & 8.055 & 50.855 & 320.795 & 5.9862 & \underline{39.97} \\
EDTalk~\cite{tan2025edtalk} & Images & 2.827 & \textbf{76.423} & \textbf{293.904} & \textbf{8.0529} & \underline{41.99} & 1.590 & \textbf{42.376} & \textbf{288.162} & \textbf{6.3569} & 29.69 \\
FLOAT~\cite{ki2024float} & Audio & \textbf{1.434} & 92.799 & 368.081 & 7.1632 & 13.21 & \textbf{0.846} & 52.933 & 365.770 & 4.9860 & 29.11 \\ 
\midrule
C-MET (Ours) & Audio & \underline{2.643} & \underline{90.804} & \underline{329.862} & 7.9996 & \textbf{55.91} & \underline{1.561} & \underline{50.028} & \underline{309.828} & \underline{6.2887} & \textbf{43.47} \\
\bottomrule
\end{tabular}
}
\vspace{-0.5em} 
\caption{\textbf{Quantitative comparison with state-of-the-art methods.} Each method is evaluated on the MEAD and CREMA-D datasets. To assess emotion editing, we input a neutral talking-face video while varying the emotion source: images (EAMM, EDTalk), label (EAT), and audio (FLOAT, ours). Best and second-best results are shown in \textbf{bold} and \underline{underline}, respectively. For emotion editing in talking-face videos, achieving higher \(\text{Acc}_{\text{emo}}\) is the primary objective, while other perceptual attributes are expected to be preserved with minimal degradation.}
\label{tab:Table01}
\vspace{-1.6em} 
\end{table*}

\subsection{Contrastive Learning on Multimodal Tokens}

To align representations from different modalities, we apply contrastive learning. Specifically, we construct the visual tokenizer \(T_{\text{v}}\) using 1D convolution layers, inspired by IP-LAP~\cite{zhong2023identity}, and the audio tokenizer \(T_{\text{a}}\) using projection layers. By leveraging the pretrained facial expression encoder \(E_v\), the visual token is then extracted as \(v = \text{Mean}(T_{v}(E_v(V_{1:T}))) \in \mathbb{R}^{d}\), where \text{Mean} denotes temporal average pooling for \(T = 5\) adjacent frames at a time. Similarly, the audio token is computed as \(a = T_{a}(E_a(A)) \in \mathbb{R}^{d}\), using the pretrained audio encoder \(E_a\).
Here, \(d\) denotes the token dimension. Inspired by the cross-modal semantic contrastive loss~\cite{mahmud2024ma}, we define the multimodal token contrastive loss as:

\begin{equation}
L_{v \rightarrow a} = -\sum_{i\in B}\log \frac{e^{(\text{sim}(v^i, a^i)/\tau)}}{e^{(\text{sim}(v^i, a^i)/\tau)}+\sum\limits_{j,i\neq j} e^{(\text{sim}(v^i, a^j)/\tau)}}  
\end{equation}
\begin{equation}
L_{a \rightarrow v} = -\sum_{i\in B}\log \frac{e^{(\text{sim}(a^i, v^i)/\tau)}}{e^{(\text{sim}(a^i, v^i)/\tau)}+\sum\limits_{j,i\neq j} e^{(\text{sim}(a^i, v^j)/\tau)}} 
\end{equation}
\begin{equation}
L_{\text{cnt}} = \frac{L_{v \rightarrow a} + L_{a \rightarrow v}}{2}    
\end{equation}

where \(\text{sim}(v^i, a^i)\) denotes the cosine similarity between visual token and audio token, \(B\) refers to the batch size, and \( \tau \) is a temperature parameter.

\begin{figure*}[h]
  \centering
  \includegraphics[width=0.9\textwidth]{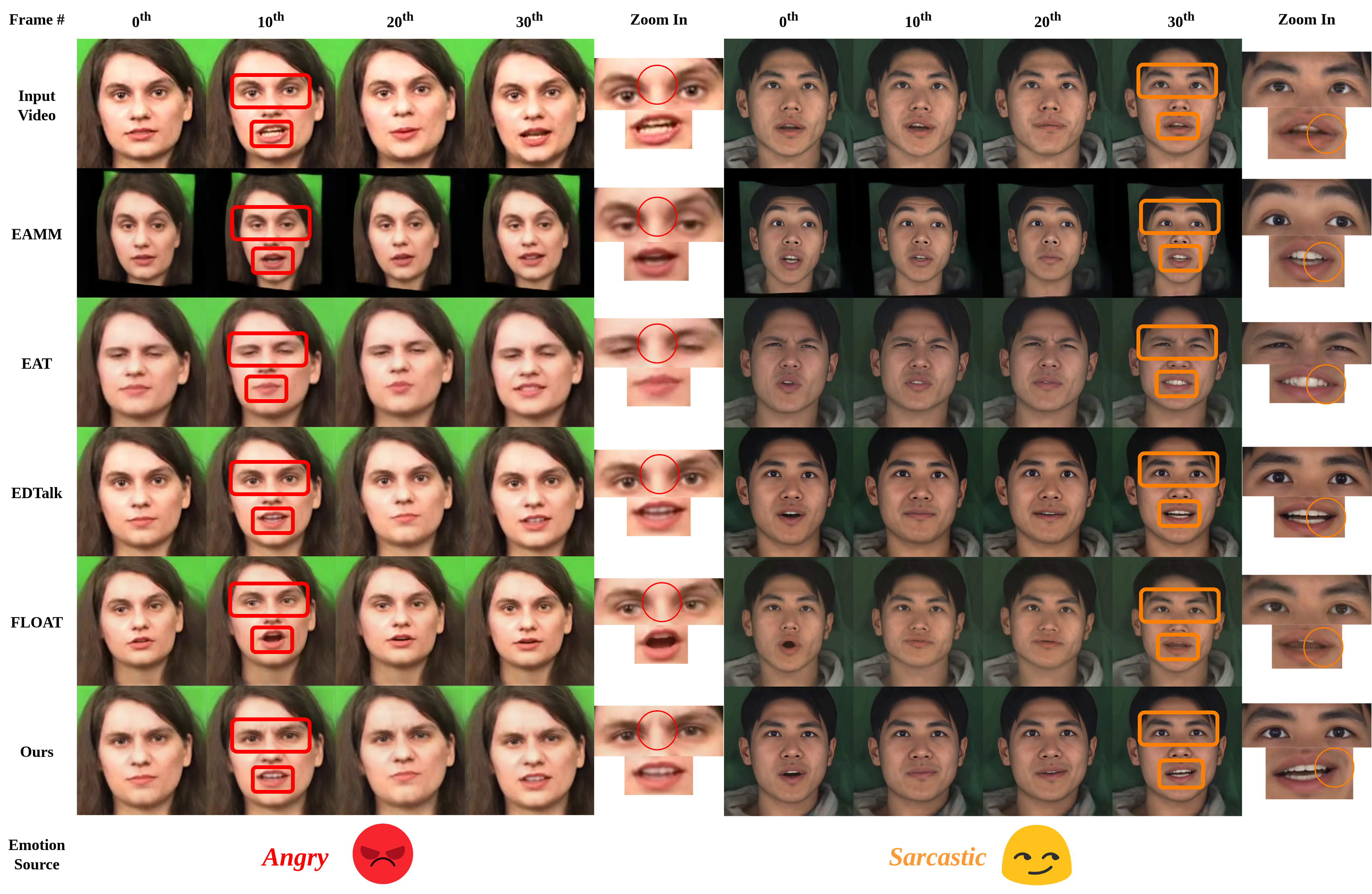}
  \vspace{-3mm} 
  \caption{Qualitative results for \textit{angry} (left) and \textit{sarcastic} (right), respectively.}
  \label{fig:qualitative}
  \vspace{-5mm} 
\end{figure*}

\subsection{Cross-Modal Emotion Transfer Learning}

To obtain an emotion semantic vector, we randomly set two different emotional states: input emotion (\(i\)) and target emotion (\(j\)).
In the audio representation space, we compute the emotion semantic vector as:
\(f^{i\rightarrow j}_a = f^j_a - f^i_a\)
where the audio embeddings \(f^i_a\) and \(f^j_a\) are encoded using the audio encoder \(E_A\). This vector is fed into transformer encoder as condition signal. In the visual representation, we define the emotion semantic vector as:
\(f^{i\rightarrow j}_{v,1:T} = f^{j}_{v,1:T} - f^{i}_{v,1:T}\)
where the visual embeddings \(f^{i}_{v,1:T}\) and \(f^{j}_{v,1:T}\) are encoded using the visual encoder \(E_V\). The vectors serve as the target of transformer encoder, and those from \textit{T} steps earlier are used as reference, denoted as \(r\).

Given the audio-visual embeddings, we construct the input tokens for the Transformer as follows:

\begin{equation}
    z_{r,t'} = f^{i\rightarrow j,t'}_v + e^{\text{pos}}_r + e^{\text{type}}_r, \quad t' = 0, 1, 2, \dots, T
\end{equation}
\begin{equation}
    z_a = f^{i\rightarrow j}_a + e^{\text{type}}_a
\end{equation}
\begin{equation}
    z_{v,t} = f^{i,t}_v + e^{\text{pos}}_v + e^{\text{type}}_v, \quad t = 1, 2, \dots, T
\end{equation}

To distinguish the embeddings derived from the three types of source signals to specify each distinct features, we introduce three learnable type embeddings: \(e^{\text{type}}_r\), \(e^{\text{type}}_a\), and \(e^{\text{type}}_v \in \mathbb{R}^d\). In addition, we apply sinusoidal positional encoding to each frame, denoted as \(e^{\text{pos}}_v \in \mathbb{R}^d\). \(z_{r,t'},z_a, z_{v,t} \in \mathbb{R}^d\) represent the transformer input tokens for the reference visual semantic vector, target speech semantic vector, and input visual embedding, respectively.

We concatenate all tokens and feed them into a stack of Transformer encoder layers~\cite{vaswani2017attention} to model both intra-modal and inter-modal dependencies.
From the final layer output, the last \(T\) visual tokens to predict the target emotion semantic vectors:

\begin{equation}
    \hat{f}^{i\rightarrow j}_{v,t} = P_{v}(TE(\{z_{r,t'}\}_{t'=0}^{T} \,\|\, \{z_a\} \,\|\, \{z_{v,t}\}_{t=1}^{T}))
\end{equation}

where \(TE\) denotes the Transformer encoder, \(||\) indicates token concatenation, and \(P_v\) denotes a projection layer to predict target visual semantic vectors.

To train the model, we minimize the mean squared error (MSE) between the predicted and target semantic vectors. Since vectors consider forward and reverse, the reconstruction loss can be summarized as:

\begin{equation}
    L_{i\rightarrow j}  = \sum_{t=1}^{T} \left\| f^{i\rightarrow j}_{v,t} - \hat{f}^{i\rightarrow j}_{v,t} \right\|_2
\end{equation}
\begin{equation}
    L_{j\rightarrow i}  = \sum_{t=1}^{T} \left\| f^{j\rightarrow i}_{v,t} - \hat{f}^{j\rightarrow i}_{v,t} \right\|_2
\end{equation}
\begin{equation}
    L_{\text{recon}}  =  L_{i\rightarrow j} + L_{j\rightarrow i}
\end{equation}

To encourage the two vectors to be opposite, we add a direction loss term: \(L_{\text{dir}}=1+\frac{\langle \hat{f}^{i\rightarrow j}_v,\hat{f}^{j\rightarrow i}_v\rangle}{||\hat{f}^{i\rightarrow j}_v|| ||\hat{f}^{j\rightarrow i}_v||}\) to the training loss. The final total loss is:
\begin{equation}
    L = L_{\text{recon}} + \lambda_{\text{cnt}} \cdot L_{\text{cnt}} + \lambda_{\text{dir}}\cdot L_{\text{dir}}
\end{equation}
where \(\lambda_{\text{cnt}}\) and \(\lambda_{\text{dir}}\) refers to the hyperparameter of \(L_{\text{cnt}}\) and \(L_{\text{dir}}\), respectively.
\section{Experiment}
\subsection{Experimental Settings}
\noindent \textbf{Implementation Details.} We adopt the audio encoder from emotion2vec+large~\cite{ma2023emotion2vec}, which extracts emotion representations across diverse tasks from a massive speech corpus in a self-supervised manner.
For the visual modality, we use the disentangled facial expression encoder and generator of EDTalk~\cite{tan2025edtalk} as the encoder and decoder, respectively.
The Transformer encoder’s multimodal token dimension \(d\) is set to 1024, and the hidden dimension is also 1024.
We use 5 frames for both reference visual semantic vectors and input embeddings.
During inference, zero padding is applied to initialize the reference, and predicted vectors are fed in an autoregressive manner.
Our model is implemented in PyTorch and trained using the AdamW optimizer on a single NVIDIA GeForce RTX 3090 GPU (24 GB).
The loss coefficients $\lambda_{\text{cnt}}$ and $\lambda_{\text{dir}}$ are set to 0.1 and 0.05, respectively.

\noindent\textbf{Dataset.}
The model is trained on the MEAD training set~\cite{wang2020mead} and evaluated on the MEAD test set and CREMA-D~\cite{cao2014crema}.
MEAD is currently the largest publicly available emotional talking audio-visual dataset.
CREMA-D includes a wide variety of speaker identities, making it suitable for assessing generalization capability.
For qualitative analysis, we also use HDTF~\cite{zhang2021flow} videos and portrait images generated by ChatGPT-4o~\cite{hurst2024gpt}.
All frames are preprocessed following EDTalk’s protocol, including face cropping and resizing to \(256 \times 256\). Audio is sampled at 16 kHz, and Mel-spectrograms are computed using a window size of 800 and hop size of 200.

\noindent\textbf{Training Data Details.}
For the speech modality, we randomly sample ten neutral and ten emotional speech clips, regardless of speaker identity or linguistic content, compute emotion semantic vectors for each pair, and average them to obtain a robust speech emotion representation. For the video modality, we similarly randomly sample ten neutral and ten emotional video clips within the same speaker identity, irrespective of head motion, and average the resulting emotion semantic vectors. This sampling-and-averaging strategy was empirically chosen to reduce noise and stabilize learning.

\noindent\textbf{Comparison Setting.}
Since our objective is to transform a neutral video into an emotional one while maintaining talking face attributes, we follow the evaluation protocol of prior emotional talking face generation works~\cite{sun2023continuously,tan2025edtalk,ki2024float}.
Video quality is measured by Fréchet Inception Distance (FID)~\cite{heusel2017gans}, temporal coherence by Fréchet Video Distance (FVD)~\cite{unterthiner2018towards}, and audio-visual synchronization by the confidence score from SyncNet~\cite{chung2016out}.
For emotional accuracy, we fine-tune Emotion-FAN~\cite{meng2019frame} on each benchmark and compute \(\text{Acc}_{\text{emo}}\). To assess computational efficiency, we report the Average Inference Time per Video (AITV), adapted from motion generation tasks~\cite{chen2023executing, dai2024motionlcm}. We compare our method with the following baselines:
(1) a label-based method (EAT~\cite{gan2023efficient});
(2) images-based methods (EAMM~\cite{ji2022eamm} and EDTalk~\cite{tan2025edtalk}, specifically the EDTalk-A, denoted simply as EDTalk); and
(3) an audio-based method (FLOAT~\cite{ki2024float}).

We evaluate two settings: \textbf{basic emotions} and \textbf{extended emotions}.
For the basic-emotion setting, we use a subset of the MEAD test set containing identical sentences spoken across eight discrete emotions to focus on expression changes from neutral to emotional states.
During emotion editing, all components except the emotion sources are fixed, using neutral audio and video as lip and pose drivers.
To test generalization to extended emotions~\cite{chen-etal-2024-emoknob} (\textit{Desire, Envy, Romance, Sarcasm, Charisma, Empathy}), we synthesize emotional speeches with Gemini TTS~\cite{team2023gemini, comanici2025gemini} for emotion sources (see supplementary for details).
Because there are no ground-truth videos for these emotions, user studies are conducted for evaluation. Audio-based methods can naturally handle speeches as emotion sources, whereas the other baselines cannot.
To reduce the domain gap and ensure fair comparison, we use emotion2vec+large to predict emotion labels and retrieve reference videos as emotion sources for the baselines.

\begin{table}[t]
\centering
\setlength{\tabcolsep}{2mm}
\begin{tabular}{ccc|c}
\toprule
\multicolumn{3}{c|}{\textbf{Loss}} & \textbf{Metric} \\
 \(L_{\text{recon}}\) & \(L_{\text{cnt}}\) & \(L_{\text{dir}}\) & $\text{Acc}_{\text{emo}}\,\uparrow$ \\
\midrule
\(\checkmark\) &  &  & 49.43 \\
\(\checkmark\) & \(\checkmark\) &  & 53.46 \\
\(\checkmark\) & \(\checkmark\) & \(\checkmark\) & \textbf{55.91} \\
\bottomrule
\end{tabular}
\vspace{-0.5em} 
\caption{We evaluate the impact of ablation on training loss.}
\vspace{-0.5em} 
\label{tab:ablation_training}
\end{table}

\begin{table}[t]
\centering
\setlength{\tabcolsep}{1.6mm} 
\renewcommand{\arraystretch}{1.05} 
\begin{tabular}{c|cc}
\toprule
\textbf{Disentanglement} &
\multicolumn{2}{c}{\textbf{Metric}} \\ 
\textbf{Network} & AITV\,$\downarrow$ & 
$\text{Acc}_{\text{emo}}\,\uparrow$ \\
\midrule
PD-FGC~\cite{wang2023progressive} & 1.247 & 33.36 \\
\textbf{w/ Ours} & \textbf{1.180} & \textbf{36.82} \\
\midrule
EDTalk~\cite{tan2025edtalk} & 2.827 & 41.99 \\
\textbf{w/ Ours} & \textbf{2.643} & \textbf{55.91} \\
\bottomrule
\end{tabular}
\vspace{-0.5em} 
\caption{Effect of integrating C-MET into disentanglement networks on MEAD.}
\label{tab:ablation_network}
\vspace{-1em} 
\end{table}

\begin{figure}[t]
  \centering
  \includegraphics[width=\linewidth]{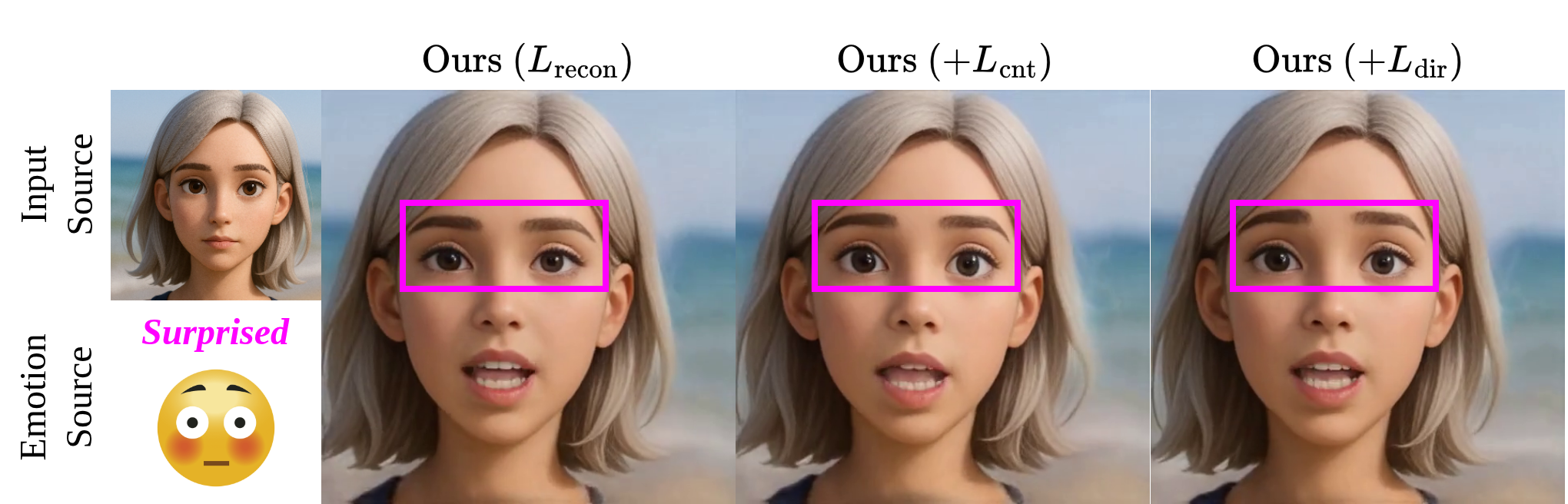}
  \vspace{-1.75em} 
  \caption{Qualitative analysis of ablation in the training loss.}
  \vspace{-1.5em} 
  \label{fig:ablation}
\end{figure}

\subsection{Experimental Results}
\noindent\textbf{Quantitative Results.}
Table~\ref{tab:Table01} presents a quantitative comparison with state-of-the-art methods in emotion editing of talking face video on the MEAD and CREMA-D.
All methods use the same neutral videos as input and differ only in the modality of the emotion source:
EAMM and EDTalk use target videos,
EAT uses target emotion labels,
while FLOAT and our method use target emotional speeches.

As shown in Table~\ref{tab:Table01}, our method achieves the highest emotion classification accuracy (\(\text{Acc}_{\text{emo}}\)) across all benchmarks, consistently outperforming state-of-the-art methods.
Although EDTalk attains slightly better scores in FID, FVD, and \(\text{Sync}_{\text{conf}}\), our method remains marginally comparable in visual quality while producing more dynamic and emotionally expressive facial motions.
This observation underscores an inherent trade-off between emotional accuracy and visual fidelity: stronger and more diverse emotional expressions often introduce larger motion and pixel deviations, resulting in minor degradation in reconstruction-based metrics.
To further assess perceptual quality beyond quantitative metrics, we conducted a user study evaluating human judgments on emotional expressiveness and visual realism. Moreover, our model achieves a lower Average Inference Time per Video (AITV), highlighting its computational efficiency compared to image-based methods that rely on heavy facial expression encoders.

\begin{figure}[t]
  \centering
  \includegraphics[width=\linewidth]{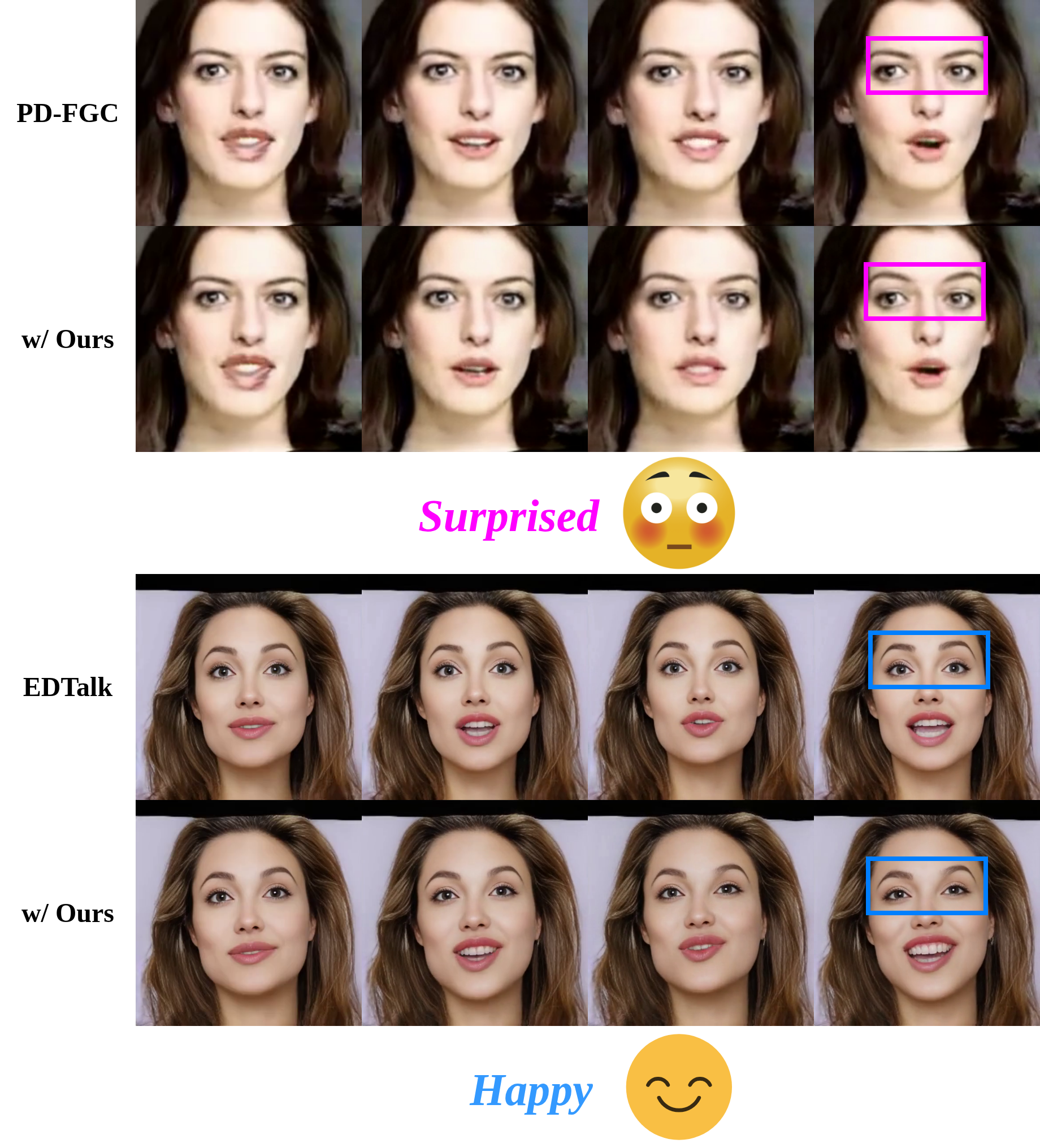}
  \vspace{-2em} 
  \caption{Qualitative analysis of integrating C-MET into disentanglement networks.}
  \vspace{-1.5em} 
  \label{fig:disentanglement}
\end{figure}

\begin{table*}[t]
\centering
\small 
\setlength{\tabcolsep}{1.4mm} 
\renewcommand{\arraystretch}{1.1} 
\begin{tabular}{l l|ccc|ccc|ccc|ccc}
\toprule
 & \textbf{Metric} 
 & \textbf{Ours} & \textbf{EAMM} & \textbf{Tie}
 & \textbf{Ours} & \textbf{EAT} & \textbf{Tie}
 & \textbf{Ours} & \textbf{EDTalk} & \textbf{Tie}
 & \textbf{Ours} & \textbf{FLOAT} & \textbf{Tie} \\ 
\midrule

\multirow{3}{*}{\textbf{\makecell{Basic\\Emotion}}}
& Emotional Expression (\%) & \textbf{77.8} & 10.4 & 11.8 & \textbf{61.6} & 21.4 & 17.1 & \textbf{42.4} & 22.0 & 35.7 & \textbf{84.5} & 14.9 & 0.6 \\
& Visual Quality (\%)       & \textbf{77.8} & 10.6 & 11.6 & \textbf{61.4} & 22.4 & 16.3 & \textbf{40.6} & 23.5 & 35.9 & \textbf{81.4} & 18.2 & 0.4 \\
& Lip Synchronization (\%)  & \textbf{71.4} & 14.5 & 14.1 & \textbf{58.2} & 24.7 & 17.1 & \textbf{40.4} & 23.3 & 36.3 & \textbf{79.0} & 20.4 & 0.6 \\
\midrule

\multirow{3}{*}{\textbf{\makecell{Extended\\Emotion}}}
& Emotional Expression (\%) & \textbf{91.0} & 6.7 & 2.2 & \textbf{80.4} & 17.1 & 2.4 & \textbf{51.2} & 36.5 & 12.2 & \textbf{86.9} & 11.8 & 1.2 \\
& Visual Quality (\%)       & \textbf{90.8} & 8.8 & 0.4 & \textbf{77.8} & 19.4 & 2.9 & \textbf{48.0} & 39.8 & 12.2 & \textbf{87.1} & 11.4 & 1.4 \\
& Lip Synchronization (\%)  & \textbf{87.6} & 9.8 & 2.7 & \textbf{78.4} & 19.4 & 2.9 & \textbf{45.3} & 39.6 & 15.1 & \textbf{86.7} & 11.8 & 1.4 \\
\bottomrule
\end{tabular}
\vspace{-0.5em} 
\caption{\textbf{User study results across basic and extended emotions.} 
We report the percentage of participants who preferred our method, a baseline, or rated them equally (tie), 
in terms of emotional expression, visual quality, and lip synchronization. 
Our method consistently outperforms all baselines across both emotion categories.}
\vspace{-1.5em} 
\label{tab:user_study}
\end{table*}



\noindent \textbf{Qualitative Results.}
Figure~\ref{fig:qualitative} presents a qualitative comparison of emotion editing results based on neutral talking face videos.
EAT produces unnatural expressions, often limited to repetitive eye closing, failing to convey coherent emotional intent.
Although both EAMM and EDTalk use target emotional videos as references, EAMM suffers from low visual fidelity, while EDTalk—despite generating sharper frames—often fails to capture the intended emotion when the reference video does not perfectly match the target expression.
FLOAT, on the other hand, cannot accurately reproduce the target emotion because it does not disentangle the neutral speech (used for lip synchronization) from the emotional speech (used for emotion source), frequently resulting in neutral or ambiguous facial outputs.

In contrast, our method generalizes across a wide range of emotions and identities by learning emotion semantic vectors that are disentangled from audio content, enabling more consistent and expressive emotion editing.
For instance, in the \textit{angry} case, our model generates dynamic frowning and eyebrow contraction that clearly convey anger.
In the \textit{sarcastic} case, our approach captures asymmetric facial nuances such as a subtle one-sided smile—an ability not exhibited by other baselines (which instead use \textit{contempt} as the closest available emotion source).
Additional qualitative results, including both basic and extended 
emotion examples, as well as confusion matrices for emotion 
consistency evaluation, are provided in the supplementary material.

\noindent \textbf{User Study.}
We conduct a user study to compare our method with baseline approaches in terms of emotional expression, visual quality, and lip synchronization. Participants are asked to choose the video that (1) best reflects the emotion conveyed by the audio, (2) exhibits higher visual quality, and (3) provides more accurate lip synchronization. As shown in Table~\ref{tab:user_study}, our method substantially outperforms all baselines for both basic and extended emotions. These results indicate that our approach more effectively edits facial expressions to match target emotions while preserving high visual fidelity and lip synchronization in human perception. (See supplementary materials for details.)

\noindent\textbf{Ablation Study.}
We conduct ablation experiments on the MEAD dataset to evaluate 
the contribution of each component in our training pipeline.
The quantitative results are summarized in 
Table~\ref{tab:ablation_training}. As illustrated in 
Figure~\ref{fig:ablation}, although using only the reconstruction 
loss provides a reasonable baseline, it is insufficient to capture 
fine-grained semantic vectors.
Introducing the contrastive loss enhances cross-modal alignment 
between audio and visual emotion representations, resulting in more 
accurate prediction.
Finally, incorporating the direction loss explicitly guides the 
model to learn emotion semantic vectors in the latent space, 
yielding the best overall performance.
Additional ablation on the choice of audio encoder is provided 
in the supplementary material.

\noindent\textbf{Generalization of Disentanglement Networks.}
As shown in Table~\ref{tab:ablation_network}, replacing the facial expression encoder with C-MET consistently improves both emotion accuracy and inference speed. This efficiency comes from replacing the heavy encoder with a more lightweight transformer-based module.
Furthermore, the results suggest that as more advanced disentanglement networks are developed, our model can seamlessly integrate with them and inherit their improvements.
As illustrated in Figure~\ref{fig:disentanglement}, our method also produces more distinctive expressions—such as a higher eyebrow lift for \textit{surprised} and more pronounced eye smiling for \textit{happy}—compared to PD-FGC and EDTalk.

\begin{figure}[t]
  \centering
  \includegraphics[width=\linewidth]{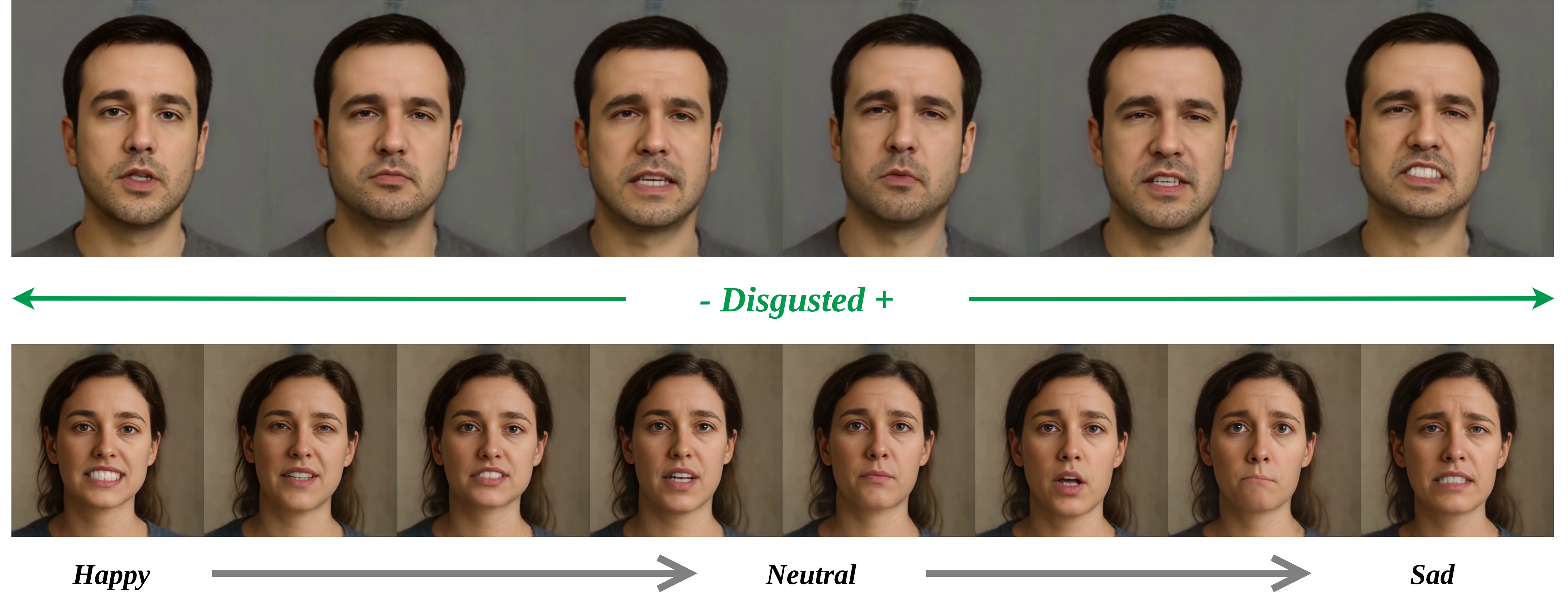}
  \vspace{-1.75em} 
  \caption{The results of continuous emotion editing.}
  \vspace{-1.5em} 
  \label{fig:continuous}
\end{figure}

\noindent\textbf{Continuous Emotion Editing.}
Our model is capable of continuous emotion editing by processing semantic vectors within short temporal windows of five frames.
During inference, C-MET sequentially applies different speech-derived emotion semantic vectors at each interval, enabling smooth and continuous facial expression transitions over time, as illustrated in Figure~\ref{fig:continuous}.
Since emotional intensity is inherently encoded in speech, our model naturally produces fine-grained facial expressions.

\section{Conclusion}

In this work, we present Cross-Modal Emotion Transfer (C-MET), a novel approach that maps emotion semantic vectors from speech to facial expressions for emotion editing in talking face videos. By learning these vectors in separate audio and visual embedding spaces, C-MET can synthesize unseen emotional expressions using expressive speech, despite being trained only on existing audio–visual datasets. The model also integrates seamlessly as a plug-and-play module into disentanglement-based generators, enhancing emotional expressiveness while reducing inference latency. Extensive experiments on MEAD and CREMA-D demonstrate that our model significantly outperforms state-of-the-art emotion editing methods in emotion accuracy while preserving visual attributes.
\paragraph*{Acknowledgment} 
This work was supported by Institute of Information \& communications Technology Planning \& Evaluation (IITP)
grant funded by the Korea government (MSIT) (No.RS-2022-II220608/2022-0-00608, Artificial intelligence research about
multimodal interactions for empathetic conversations with humans, No.IITP-2026-RS-2024-00360227, Leading Generative AI Human Resources Development, No.RS-2025-25442824, AI Star Fellowship Program(Ulsan National Institute of Science and Technology, \& No.RS-2020-II201336, Artificial Intelligence graduate school support(UNIST)).
{
    \small
    \bibliographystyle{ieeenat_fullname}
    \bibliography{main}

@String(ICCV= {Int. Conf. Comput. Vis.})

@String(ECCV= {Eur. Conf. Comput. Vis.})

@String(TOG= {ACM Trans. Graph.})

@String(ICIP = {IEEE Int. Conf. Image Process.})

@String(IJCAI = {IJCAI})

@String(AAAI = {AAAI})

@String(ICCV  = {ICCV})

@String(ECCV  = {ECCV})

@String(TOG   = {ACM TOG})

@String(ICIP  = {ICIP})

@inproceedings{chen-etal-2024-emoknob,
    title = "{E}mo{K}nob: Enhance Voice Cloning with Fine-Grained Emotion Control",
    author = "Chen, Haozhe  and
      Chen, Run  and
      Hirschberg, Julia",
    editor = "Al-Onaizan, Yaser  and
      Bansal, Mohit  and
      Chen, Yun-Nung",
    booktitle = "Proceedings of the 2024 Conference on Empirical Methods in Natural Language Processing",
    month = nov,
    year = "2024",
    address = "Miami, Florida, USA",
    publisher = "Association for Computational Linguistics",
    url = "https://aclanthology.org/2024.emnlp-main.466/",
    doi = "10.18653/v1/2024.emnlp-main.466",
    pages = "8170--8180"
}

@inproceedings{gan2023efficient,
  title={Efficient emotional adaptation for audio-driven talking-head generation},
  author={Gan, Yuan and Yang, Zongxin and Yue, Xihang and Sun, Lingyun and Yang, Yi},
  booktitle={Proceedings of the IEEE/CVF International Conference on Computer Vision},
  pages={22634--22645},
  year={2023}
}

@inproceedings{tan2025edtalk,
  title={Edtalk: Efficient disentanglement for emotional talking head synthesis},
  author={Tan, Shuai and Ji, Bin and Bi, Mengxiao and Pan, Ye},
  booktitle={European Conference on Computer Vision},
  pages={398--416},
  year={2024},
  organization={Springer}
}

@inproceedings{wang2020mead,
  title={Mead: A large-scale audio-visual dataset for emotional talking-face generation},
  author={Wang, Kaisiyuan and Wu, Qianyi and Song, Linsen and Yang, Zhuoqian and Wu, Wayne and Qian, Chen and He, Ran and Qiao, Yu and Loy, Chen Change},
  booktitle={European conference on computer vision},
  pages={700--717},
  year={2020},
  organization={Springer}
}

@article{pataranutaporn2021ai,
  title={AI-generated characters for supporting personalized learning and well-being},
  author={Pataranutaporn, Pat and Danry, Valdemar and Leong, Joanne and Punpongsanon, Parinya and Novy, Dan and Maes, Pattie and Sra, Misha},
  journal={Nature Machine Intelligence},
  volume={3},
  number={12},
  pages={1013--1022},
  year={2021},
  publisher={Nature Publishing Group UK London}
}

@inproceedings{goyal2023emotionally,
  title={Emotionally enhanced talking face generation},
  author={Goyal, Sahil and Bhagat, Sarthak and Uppal, Shagun and Jangra, Hitkul and Yu, Yi and Yin, Yifang and Shah, Rajiv Ratn},
  booktitle={Proceedings of the 1st International Workshop on Multimedia Content Generation and Evaluation: New Methods and Practice},
  pages={81--90},
  year={2023}
}

@inproceedings{wang2023progressive,
  title={Progressive disentangled representation learning for fine-grained controllable talking head synthesis},
  author={Wang, Duomin and Deng, Yu and Yin, Zixin and Shum, Heung-Yeung and Wang, Baoyuan},
  booktitle={Proceedings of the IEEE/CVF Conference on Computer Vision and Pattern Recognition},
  pages={17979--17989},
  year={2023}
}

@inproceedings{zhong2023identity,
  title={Identity-preserving talking face generation with landmark and appearance priors},
  author={Zhong, Weizhi and Fang, Chaowei and Cai, Yinqi and Wei, Pengxu and Zhao, Gangming and Lin, Liang and Li, Guanbin},
  booktitle={Proceedings of the IEEE/CVF Conference on Computer Vision and Pattern Recognition},
  pages={9729--9738},
  year={2023}
}

@inproceedings{ji2022eamm,
  title={Eamm: One-shot emotional talking face via audio-based emotion-aware motion model},
  author={Ji, Xinya and Zhou, Hang and Wang, Kaisiyuan and Wu, Qianyi and Wu, Wayne and Xu, Feng and Cao, Xun},
  booktitle={ACM SIGGRAPH 2022 conference proceedings},
  pages={1--10},
  year={2022}
}

@inproceedings{tan2024style2talker,
  title={Style2Talker: High-Resolution Talking Head Generation with Emotion Style and Art Style},
  author={Tan, Shuai and Ji, Bin and Pan, Ye},
  booktitle={Proceedings of the AAAI Conference on Artificial Intelligence},
  volume={38},
  number={5},
  pages={5079--5087},
  year={2024}
}

@inproceedings{radford2021learning,
  title={Learning transferable visual models from natural language supervision},
  author={Radford, Alec and Kim, Jong Wook and Hallacy, Chris and Ramesh, Aditya and Goh, Gabriel and Agarwal, Sandhini and Sastry, Girish and Askell, Amanda and Mishkin, Pamela and Clark, Jack and others},
  booktitle={International conference on machine learning},
  pages={8748--8763},
  year={2021},
  organization={PmLR}
}

@inproceedings{mahmud2024ma,
  title={Ma-avt: Modality alignment for parameter-efficient audio-visual transformers},
  author={Mahmud, Tanvir and Mo, Shentong and Tian, Yapeng and Marculescu, Diana},
  booktitle={Proceedings of the IEEE/CVF Conference on Computer Vision and Pattern Recognition},
  pages={7996--8005},
  year={2024}
}

@article{vaswani2017attention,
  title={Attention is all you need},
  author={Vaswani, Ashish and Shazeer, Noam and Parmar, Niki and Uszkoreit, Jakob and Jones, Llion and Gomez, Aidan N and Kaiser, {\L}ukasz and Polosukhin, Illia},
  journal={Advances in neural information processing systems},
  volume={30},
  year={2017}
}

@article{sun2023continuously,
  title={Continuously controllable facial expression editing in talking face videos},
  author={Sun, Zhiyao and Wen, Yu-Hui and Lv, Tian and Sun, Yanan and Zhang, Ziyang and Wang, Yaoyuan and Liu, Yong-Jin},
  journal={IEEE Transactions on Affective Computing},
  volume={15},
  number={3},
  pages={1400--1413},
  year={2023},
  publisher={IEEE}
}

@article{ma2023emotion2vec,
  title={emotion2vec: Self-Supervised Pre-Training for Speech Emotion Representation},
  author={Ma, Ziyang and Zheng, Zhisheng and Ye, Jiaxin and Li, Jinchao and Gao, Zhifu and Zhang, Shiliang and Chen, Xie},
  journal={Proc. ACL 2024 Findings},
  year={2024}
}

@inproceedings{meng2019frame,
  title={Frame attention networks for facial expression recognition in videos},
  author={Meng, Debin and Peng, Xiaojiang and Wang, Kai and Qiao, Yu},
  booktitle={2019 IEEE international conference on image processing (ICIP)},
  pages={3866--3870},
  year={2019},
  organization={IEEE}
}

@article{poria2018meld,
  title={Meld: A multimodal multi-party dataset for emotion recognition in conversations},
  author={Poria, Soujanya and Hazarika, Devamanyu and Majumder, Navonil and Naik, Gautam and Cambria, Erik and Mihalcea, Rada},
  journal={arXiv preprint arXiv:1810.02508},
  year={2018}
}

@inproceedings{yu2023talking,
  title={Talking head generation with probabilistic audio-to-visual diffusion priors},
  author={Yu, Zhentao and Yin, Zixin and Zhou, Deyu and Wang, Duomin and Wong, Finn and Wang, Baoyuan},
  booktitle={Proceedings of the IEEE/CVF International Conference on Computer Vision},
  pages={7645--7655},
  year={2023}
}

@article{heusel2017gans,
  title={Gans trained by a two time-scale update rule converge to a local nash equilibrium},
  author={Heusel, Martin and Ramsauer, Hubert and Unterthiner, Thomas and Nessler, Bernhard and Hochreiter, Sepp},
  journal={Advances in neural information processing systems},
  volume={30},
  year={2017}
}

@inproceedings{chung2016out,
  title={Out of time: automated lip sync in the wild},
  author={Chung, Joon Son and Zisserman, Andrew},
  booktitle={Asian conference on computer vision},
  pages={251--263},
  year={2016},
  organization={Springer}
}

@article{cao2014crema,
  title={Crema-d: Crowd-sourced emotional multimodal actors dataset},
  author={Cao, Houwei and Cooper, David G and Keutmann, Michael K and Gur, Ruben C and Nenkova, Ani and Verma, Ragini},
  journal={IEEE transactions on affective computing},
  volume={5},
  number={4},
  pages={377--390},
  year={2014},
  publisher={IEEE}
}

@article{zhou2020makelttalk,
  title={Makelttalk: speaker-aware talking-head animation},
  author={Zhou, Yang and Han, Xintong and Shechtman, Eli and Echevarria, Jose and Kalogerakis, Evangelos and Li, Dingzeyu},
  journal={ACM Transactions On Graphics (TOG)},
  volume={39},
  number={6},
  pages={1--15},
  year={2020},
  publisher={ACM New York, NY, USA}
}

@inproceedings{chen2019hierarchical,
  title={Hierarchical cross-modal talking face generation with dynamic pixel-wise loss},
  author={Chen, Lele and Maddox, Ross K and Duan, Zhiyao and Xu, Chenliang},
  booktitle={Proceedings of the IEEE/CVF conference on computer vision and pattern recognition},
  pages={7832--7841},
  year={2019}
}

@inproceedings{das2020speech,
  title={Speech-driven facial animation using cascaded gans for learning of motion and texture},
  author={Das, Dipanjan and Biswas, Sandika and Sinha, Sanjana and Bhowmick, Brojeshwar},
  booktitle={Computer Vision--ECCV 2020: 16th European Conference, Glasgow, UK, August 23--28, 2020, Proceedings, Part XXX 16},
  pages={408--424},
  year={2020},
  organization={Springer}
}

@inproceedings{zakharov2019few,
  title={Few-shot adversarial learning of realistic neural talking head models},
  author={Zakharov, Egor and Shysheya, Aliaksandra and Burkov, Egor and Lempitsky, Victor},
  booktitle={Proceedings of the IEEE/CVF international conference on computer vision},
  pages={9459--9468},
  year={2019}
}

@inproceedings{wang2021audio2head,
  title={Audio2Head: Audio-driven One-shot Talking-head Generation with Natural Head Motion},
  author={Wang, S and Li, L and Ding, Y and Fan, C and Yu, X},
  booktitle={International Joint Conference on Artificial Intelligence},
  year={2021},
  organization={IJCAI}
}

@inproceedings{wang2022one,
  title={One-shot talking face generation from single-speaker audio-visual correlation learning},
  author={Wang, Suzhen and Li, Lincheng and Ding, Yu and Yu, Xin},
  booktitle={Proceedings of the AAAI Conference on Artificial Intelligence},
  volume={36},
  number={3},
  pages={2531--2539},
  year={2022}
}

@inproceedings{chen2020talking,
  title={Talking-head generation with rhythmic head motion},
  author={Chen, Lele and Cui, Guofeng and Liu, Celong and Li, Zhong and Kou, Ziyi and Xu, Yi and Xu, Chenliang},
  booktitle={European Conference on Computer Vision},
  pages={35--51},
  year={2020},
  organization={Springer}
}

@inproceedings{yang2022face2face,
  title={Face2Face $\rho$: Real-Time High-Resolution One-Shot Face Reenactment},
  author={Yang, Kewei and Chen, Kang and Guo, Daoliang and Zhang, Song-Hai and Guo, Yuan-Chen and Zhang, Weidong},
  booktitle={European conference on computer vision},
  pages={55--71},
  year={2022},
  organization={Springer}
}

@article{hernandez2023affective,
  title={Affective conversational agents: understanding expectations and personal influences},
  author={Hernandez, Javier and Suh, Jina and Amores, Judith and Rowan, Kael and Ramos, Gonzalo and Czerwinski, Mary},
  journal={arXiv preprint arXiv:2310.12459},
  year={2023}
}

@article{rings2024empathy,
  title={Empathy in Virtual Agents: How Emotional Expressions can Influence User Perception},
  author={Rings, Sebastian and Schmidt, Susanne and Jan{\ss}en, Julia and Lehmann-Willenbrock, Nale and Steinicke, Frank and Hasegawa, S and Sakata, N and Sundstedt, V},
  journal={ICAT-EGVE},
  year={2024}
}

@article{saffaryazdi2025empathetic,
  title={Empathetic Conversational Agents: Utilizing Neural and Physiological Signals for Enhanced Empathetic Interactions},
  author={Saffaryazdi, Nastaran and Gunasekaran, Tamil Selvan and Loveys, Kate and Broadbent, Elizabeth and Billinghurst, Mark},
  journal={International Journal of Human--Computer Interaction},
  pages={1--25},
  year={2025},
  publisher={Taylor \& Francis}
}

@article{ki2024float,
  title={Float: Generative motion latent flow matching for audio-driven talking portrait},
  author={Ki, Taekyung and Min, Dongchan and Chae, Gyeongsu},
  journal={arXiv preprint arXiv:2412.01064},
  year={2024}
}

@inproceedings{wang2024eat,
  title={Eat-face: Emotion-controllable audio-driven talking face generation via diffusion model},
  author={Wang, Haodi and Jia, Xiaojun and Cao, Xiaochun},
  booktitle={2024 IEEE 18th International Conference on Automatic Face and Gesture Recognition (FG)},
  pages={1--10},
  year={2024},
  organization={IEEE}
}

@inproceedings{lin2025mvportrait,
  title={Mvportrait: Text-guided motion and emotion control for multi-view vivid portrait animation},
  author={Lin, Yukang and Fung, Hokit and Xu, Jianjin and Ren, Zeping and Lau, Adela SM and Yin, Guosheng and Li, Xiu},
  booktitle={Proceedings of the Computer Vision and Pattern Recognition Conference},
  pages={26242--26252},
  year={2025}
}

@inproceedings{tan2023emmn,
  title={Emmn: Emotional motion memory network for audio-driven emotional talking face generation},
  author={Tan, Shuai and Ji, Bin and Pan, Ye},
  booktitle={Proceedings of the IEEE/CVF International Conference on Computer Vision},
  pages={22146--22156},
  year={2023}
}

@inproceedings{shen2024emotalker,
  title={EmoTalker: Audio Driven Emotion Aware Talking Head Generation},
  author={Shen, Xiaoqian and Khan, Faizan Farooq and Elhoseiny, Mohamed},
  booktitle={Proceedings of the Asian Conference on Computer Vision},
  pages={1900--1917},
  year={2024}
}

@inproceedings{dai2025emohuman,
  title={EmoHuman: Fine-Grained Emotion-Controlled Talking Head Generation via Audio-Text Multimodal Detangling},
  author={Dai, Qifeng and Feng, Huidong and Cui, Wendi and Cai, Xinqi and Zheng, Yinglin and Zeng, Ming},
  booktitle={Proceedings of the 2025 International Conference on Multimedia Retrieval},
  pages={145--154},
  year={2025}
}

@inproceedings{wang2025emotivetalk,
  title={Emotivetalk: Expressive talking head generation through audio information decoupling and emotional video diffusion},
  author={Wang, Haotian and Weng, Yuzhe and Li, Yueyan and Guo, Zilu and Du, Jun and Niu, Shutong and Ma, Jiefeng and He, Shan and Wu, Xiaoyan and Hu, Qiming and others},
  booktitle={Proceedings of the Computer Vision and Pattern Recognition Conference},
  pages={26212--26221},
  year={2025}
}

@article{shen2025emohead,
  title={EmoHead: Emotional Talking Head via Manipulating Semantic Expression Parameters},
  author={Shen, Xuli and Cai, Hua and Yu, Dingding and Shen, Weilin and Xu, Qing and Xue, Xiangyang},
  journal={arXiv preprint arXiv:2503.19416},
  year={2025}
}

@inproceedings{liu2025moee,
  title={Moee: Mixture of emotion experts for audio-driven portrait animation},
  author={Liu, Huaize and Sun, Wenzhang and Di, Donglin and Sun, Shibo and Yang, Jiahui and Zou, Changqing and Bao, Hujun},
  booktitle={Proceedings of the Computer Vision and Pattern Recognition Conference},
  pages={26222--26231},
  year={2025}
}

@inproceedings{tian2024emo,
  title={Emo: Emote portrait alive generating expressive portrait videos with audio2video diffusion model under weak conditions},
  author={Tian, Linrui and Wang, Qi and Zhang, Bang and Bo, Liefeng},
  booktitle={European Conference on Computer Vision},
  pages={244--260},
  year={2024},
  organization={Springer}
}

@inproceedings{sun2024fg,
  title={FG-EmoTalk: Talking head video generation with fine-grained controllable facial expressions},
  author={Sun, Zhaoxu and Xuan, Yuze and Liu, Fang and Xiang, Yang},
  booktitle={Proceedings of the AAAI Conference on Artificial Intelligence},
  volume={38},
  number={5},
  pages={5043--5051},
  year={2024}
}

@article{unterthiner2018towards,
  title={Towards accurate generative models of video: A new metric \& challenges},
  author={Unterthiner, Thomas and Van Steenkiste, Sjoerd and Kurach, Karol and Marinier, Raphael and Michalski, Marcin and Gelly, Sylvain},
  journal={arXiv preprint arXiv:1812.01717},
  year={2018}
}

@inproceedings{zhang2021flow,
  title={Flow-Guided One-Shot Talking Face Generation With a High-Resolution Audio-Visual Dataset},
  author={Zhang, Zhimeng and Li, Lincheng and Ding, Yu and Fan, Changjie},
  booktitle={Proceedings of the IEEE/CVF Conference on Computer Vision and Pattern Recognition},
  pages={3661--3670},
  year={2021}
}

@article{hurst2024gpt,
  title={Gpt-4o system card},
  author={Hurst, Aaron and Lerer, Adam and Goucher, Adam P and Perelman, Adam and Ramesh, Aditya and Clark, Aidan and Ostrow, AJ and Welihinda, Akila and Hayes, Alan and Radford, Alec and others},
  journal={arXiv preprint arXiv:2410.21276},
  year={2024}
}

@inproceedings{prajwal2020lip,
  title={A lip sync expert is all you need for speech to lip generation in the wild},
  author={Prajwal, KR and Mukhopadhyay, Rudrabha and Namboodiri, Vinay P and Jawahar, CV},
  booktitle={Proceedings of the 28th ACM international conference on multimedia},
  pages={484--492},
  year={2020}
}

@incollection{bregler2023video,
  title={Video rewrite: Driving visual speech with audio},
  author={Bregler, Christoph and Covell, Michele and Slaney, Malcolm},
  booktitle={Seminal Graphics Papers: Pushing the Boundaries, Volume 2},
  pages={715--722},
  year={2023}
}

@inproceedings{liu2022semantic,
  title={Semantic-aware implicit neural audio-driven video portrait generation},
  author={Liu, Xian and Xu, Yinghao and Wu, Qianyi and Zhou, Hang and Wu, Wayne and Zhou, Bolei},
  booktitle={European conference on computer vision},
  pages={106--125},
  year={2022},
  organization={Springer}
}

@inproceedings{yin2022styleheat,
  title={Styleheat: One-shot high-resolution editable talking face generation via pre-trained stylegan},
  author={Yin, Fei and Zhang, Yong and Cun, Xiaodong and Cao, Mingdeng and Fan, Yanbo and Wang, Xuan and Bai, Qingyan and Wu, Baoyuan and Wang, Jue and Yang, Yujiu},
  booktitle={European conference on computer vision},
  pages={85--101},
  year={2022},
  organization={Springer}
}

@inproceedings{chen2018lip,
  title={Lip movements generation at a glance},
  author={Chen, Lele and Li, Zhiheng and Maddox, Ross K and Duan, Zhiyao and Xu, Chenliang},
  booktitle={Proceedings of the European conference on computer vision (ECCV)},
  pages={520--535},
  year={2018}
}

@inproceedings{chen2023vast,
  title={VAST: vivify your talking avatar via zero-shot expressive facial style transfer},
  author={Chen, Liyang and Wu, Zhiyong and Li, Runnan and Bao, Weihong and Ling, Jun and Tan, Xu and Zhao, Sheng},
  booktitle={Proceedings of the IEEE/CVF International Conference on Computer Vision},
  pages={2977--2987},
  year={2023}
}

@inproceedings{shen2022learning,
  title={Learning dynamic facial radiance fields for few-shot talking head synthesis},
  author={Shen, Shuai and Li, Wanhua and Zhu, Zheng and Duan, Yueqi and Zhou, Jie and Lu, Jiwen},
  booktitle={European conference on computer vision},
  pages={666--682},
  year={2022},
  organization={Springer}
}

@inproceedings{shen2023difftalk,
  title={Difftalk: Crafting diffusion models for generalized audio-driven portraits animation},
  author={Shen, Shuai and Zhao, Wenliang and Meng, Zibin and Li, Wanhua and Zhu, Zheng and Zhou, Jie and Lu, Jiwen},
  booktitle={Proceedings of the IEEE/CVF conference on computer vision and pattern recognition},
  pages={1982--1991},
  year={2023}
}

@inproceedings{thies2020neural,
  title={Neural voice puppetry: Audio-driven facial reenactment},
  author={Thies, Justus and Elgharib, Mohamed and Tewari, Ayush and Theobalt, Christian and Nie{\ss}ner, Matthias},
  booktitle={European conference on computer vision},
  pages={716--731},
  year={2020},
  organization={Springer}
}

@inproceedings{wang2023lipformer,
  title={Lipformer: High-fidelity and generalizable talking face generation with a pre-learned facial codebook},
  author={Wang, Jiayu and Zhao, Kang and Zhang, Shiwei and Zhang, Yingya and Shen, Yujun and Zhao, Deli and Zhou, Jingren},
  booktitle={Proceedings of the IEEE/CVF Conference on Computer Vision and Pattern Recognition},
  pages={13844--13853},
  year={2023}
}

@inproceedings{zhou2019talking,
  title={Talking face generation by adversarially disentangled audio-visual representation},
  author={Zhou, Hang and Liu, Yu and Liu, Ziwei and Luo, Ping and Wang, Xiaogang},
  booktitle={Proceedings of the AAAI conference on artificial intelligence},
  volume={33},
  number={01},
  pages={9299--9306},
  year={2019}
}

@inproceedings{zhou2021pose,
  title={Pose-controllable talking face generation by implicitly modularized audio-visual representation},
  author={Zhou, Hang and Sun, Yasheng and Wu, Wayne and Loy, Chen Change and Wang, Xiaogang and Liu, Ziwei},
  booktitle={Proceedings of the IEEE/CVF conference on computer vision and pattern recognition},
  pages={4176--4186},
  year={2021}
}

@inproceedings{pang2023dpe,
  title={Dpe: Disentanglement of pose and expression for general video portrait editing},
  author={Pang, Youxin and Zhang, Yong and Quan, Weize and Fan, Yanbo and Cun, Xiaodong and Shan, Ying and Yan, Dong-ming},
  booktitle={Proceedings of the IEEE/CVF Conference on Computer Vision and Pattern Recognition},
  pages={427--436},
  year={2023}
}

@inproceedings{liang2022expressive,
  title={Expressive talking head generation with granular audio-visual control},
  author={Liang, Borong and Pan, Yan and Guo, Zhizhi and Zhou, Hang and Hong, Zhibin and Han, Xiaoguang and Han, Junyu and Liu, Jingtuo and Ding, Errui and Wang, Jingdong},
  booktitle={Proceedings of the IEEE/CVF conference on computer vision and pattern recognition},
  pages={3387--3396},
  year={2022}
}

@inproceedings{peng2023emotalk,
  title={Emotalk: Speech-driven emotional disentanglement for 3d face animation},
  author={Peng, Ziqiao and Wu, Haoyu and Song, Zhenbo and Xu, Hao and Zhu, Xiangyu and He, Jun and Liu, Hongyan and Fan, Zhaoxin},
  booktitle={Proceedings of the IEEE/CVF international conference on computer vision},
  pages={20687--20697},
  year={2023}
}

@inproceedings{ma2023styletalk,
  title={Styletalk: One-shot talking head generation with controllable speaking styles},
  author={Ma, Yifeng and Wang, Suzhen and Hu, Zhipeng and Fan, Changjie and Lv, Tangjie and Ding, Yu and Deng, Zhidong and Yu, Xin},
  booktitle={Proceedings of the AAAI conference on artificial intelligence},
  volume={37},
  number={2},
  pages={1896--1904},
  year={2023}
}

@article{ekman1992argument,
  title={An argument for basic emotions},
  author={Ekman, Paul},
  journal={Cognition \& emotion},
  volume={6},
  number={3-4},
  pages={169--200},
  year={1992},
  publisher={Taylor \& Francis}
}

@inproceedings{chen2025echomimic,
  title={Echomimic: Lifelike audio-driven portrait animations through editable landmark conditions},
  author={Chen, Zhiyuan and Cao, Jiajiong and Chen, Zhiquan and Li, Yuming and Ma, Chenguang},
  booktitle={Proceedings of the AAAI Conference on Artificial Intelligence},
  volume={39},
  number={3},
  pages={2403--2410},
  year={2025}
}

@article{xu2024vasa,
  title={Vasa-1: Lifelike audio-driven talking faces generated in real time},
  author={Xu, Sicheng and Chen, Guojun and Guo, Yu-Xiao and Yang, Jiaolong and Li, Chong and Zang, Zhenyu and Zhang, Yizhong and Tong, Xin and Guo, Baining},
  journal={Advances in Neural Information Processing Systems},
  volume={37},
  pages={660--684},
  year={2024}
}

@inproceedings{dai2024motionlcm,
  title={Motionlcm: Real-time controllable motion generation via latent consistency model},
  author={Dai, Wenxun and Chen, Ling-Hao and Wang, Jingbo and Liu, Jinpeng and Dai, Bo and Tang, Yansong},
  booktitle={European Conference on Computer Vision},
  pages={390--408},
  year={2024},
  organization={Springer}
}

@article{comanici2025gemini,
  title={Gemini 2.5: Pushing the frontier with advanced reasoning, multimodality, long context, and next generation agentic capabilities},
  author={Comanici, Gheorghe and Bieber, Eric and Schaekermann, Mike and Pasupat, Ice and Sachdeva, Noveen and Dhillon, Inderjit and Blistein, Marcel and Ram, Ori and Zhang, Dan and Rosen, Evan and others},
  journal={arXiv preprint arXiv:2507.06261},
  year={2025}
}

@article{team2023gemini,
  title={Gemini: a family of highly capable multimodal models},
  author={Team, Gemini and Anil, Rohan and Borgeaud, Sebastian and Alayrac, Jean-Baptiste and Yu, Jiahui and Soricut, Radu and Schalkwyk, Johan and Dai, Andrew M and Hauth, Anja and Millican, Katie and others},
  journal={arXiv preprint arXiv:2312.11805},
  year={2023}
}

@inproceedings{chen2023executing,
  title={Executing your commands via motion diffusion in latent space},
  author={Chen, Xin and Jiang, Biao and Liu, Wen and Huang, Zilong and Fu, Bin and Chen, Tao and Yu, Gang},
  booktitle={Proceedings of the IEEE/CVF conference on computer vision and pattern recognition},
  pages={18000--18010},
  year={2023}
}

@article{lotfian2017building,
author = {Lotfian, Reza and Busso, Carlos},
year = {2017},
month = {08},
pages = {1-1},
title = {Building Naturalistic Emotionally Balanced Speech Corpus by Retrieving Emotional Speech From Existing Podcast Recordings},
volume = {PP},
journal = {IEEE Transactions on Affective Computing},
doi = {10.1109/TAFFC.2017.2736999}
}

@article{busso2008iemocap,
author = {Busso, Carlos and Bulut, Murtaza and Lee, Chi-Chun and Kazemzadeh, Abe and Mower Provost, Emily and Kim, Samuel and Chang, Jeannette and Lee, Sungbok and Narayanan, Shrikanth},
year = {2008},
month = {12},
pages = {335-359},
title = {IEMOCAP: Interactive emotional dyadic motion capture database},
volume = {42},
journal = {Language Resources and Evaluation},
doi = {10.1007/s10579-008-9076-6}
}

@inproceedings{gururani2023space,
  title={Space: Speech-driven portrait animation with controllable expression},
  author={Gururani et al.},
  booktitle={ICCV},
  year={2023}
}

@article{xu2025qwen2,
  title={Qwen2. 5-omni technical report},
  author={Xu et al.},
  journal={arXiv},
  year={2025}
}
}

\clearpage
\setcounter{section}{0}
\renewcommand{\thesection}{\Alph{section}}
\maketitlesupplementary

In this supplementary material, we first describe how to generate expressive speeches by utilizing a large generative model. Next, we provide more visualization results. In addition, we introduce the additional experimental results (impact of speech-shot, emotion consistency evaluation, ablation study on audio encoder and full metrics of ablation study). Finally, we present human evaluation templates and limitations.

\section{Expressive Text-to-Speech}

To synthesize expressive speech for the extended emotion categories, we utilize the Gemini 2.5 Flash~\cite{comanici2025gemini} TTS framework.
For each target emotion, the language model first generates a sentence that naturally conveys the intended affective nuance.
We then query the model again to select an appropriate voice identity from a predefined set of expressive voice styles.
The selected voice identity is injected into the TTS generation pipeline through the \texttt{voice\_config} parameter of the API.
Finally, the speech waveform is synthesized using the instruction “Say with \{\texttt{emotion}\} voice: \{\texttt{sentence}\}”, conditioned jointly on the textual prompt and the chosen voice configuration.
This procedure allows us to produce consistent and controllable expressive speech samples across six extended emotions~\cite{chen-etal-2024-emoknob}.

\section{More Visualization Results}

\begin{figure}[t]
\centering
\includegraphics[width=\linewidth]{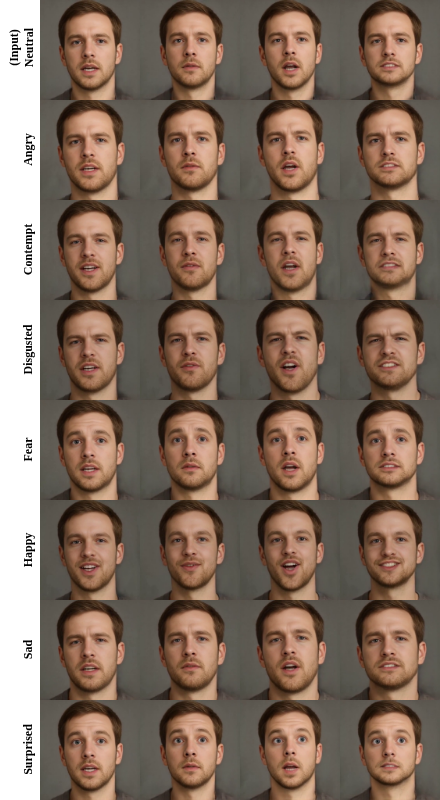}
\caption{Emotion editing results on basic emotions.}
\label{fig:supp_basic_emotion}
\end{figure}

\begin{figure}[t]
\centering
\includegraphics[width=\linewidth]{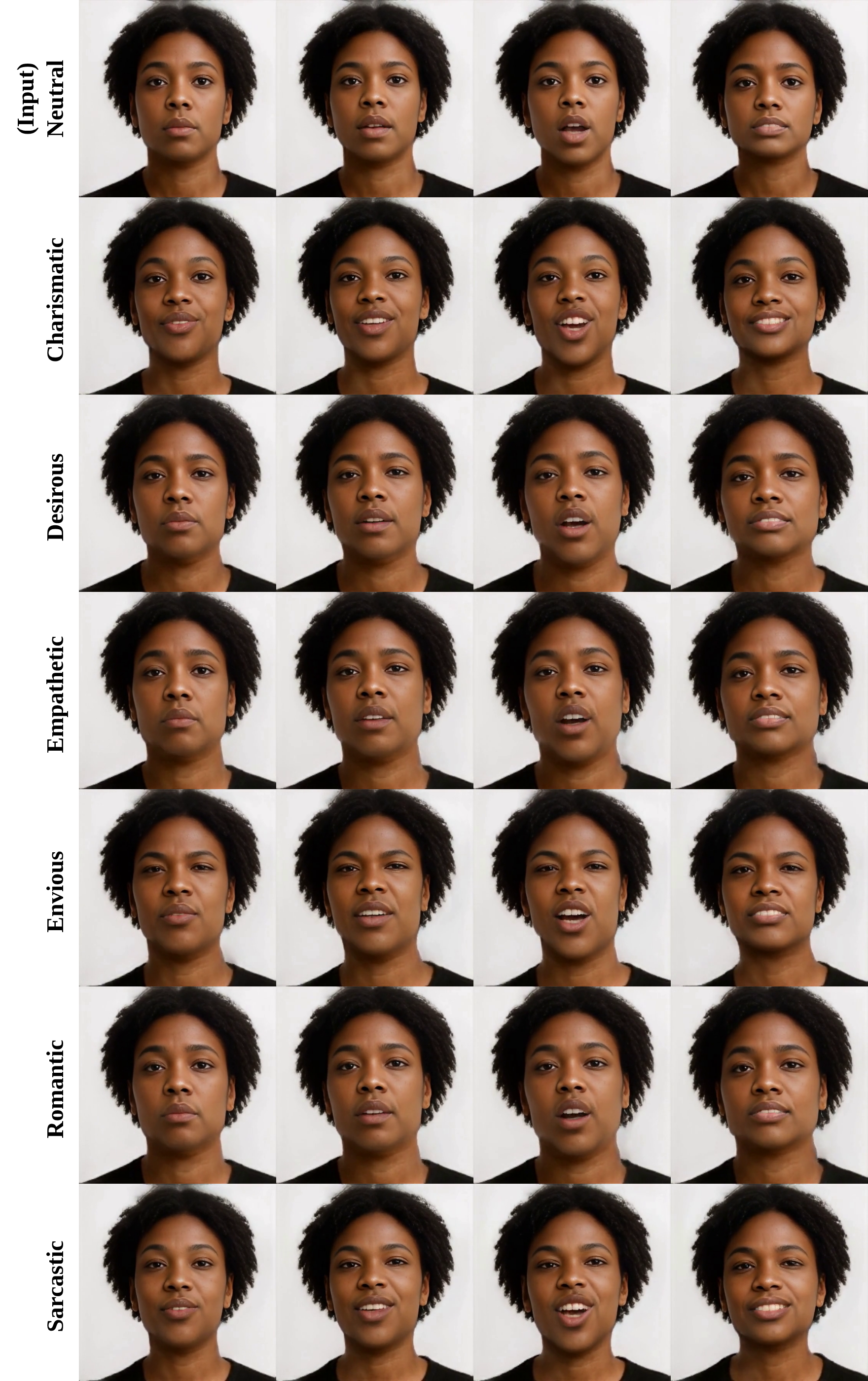}
\caption{Emotion editing results on extended emotions.}
\label{fig:supp_extended_emotion}
\end{figure}

To further support the findings presented in the main paper, we include additional qualitative examples in this section. Figures~\ref{fig:supp_basic_emotion} and~\ref{fig:supp_extended_emotion} provide more visualization results of our method across various emotions.
Interactive playback of the sample videos is available on our project 
page: \url{https://chanhyeok-choi.github.io/C-MET/}.

\section{Additional Experimental Results}

\begin{figure}[t]
\centering
\includegraphics[width=\linewidth]{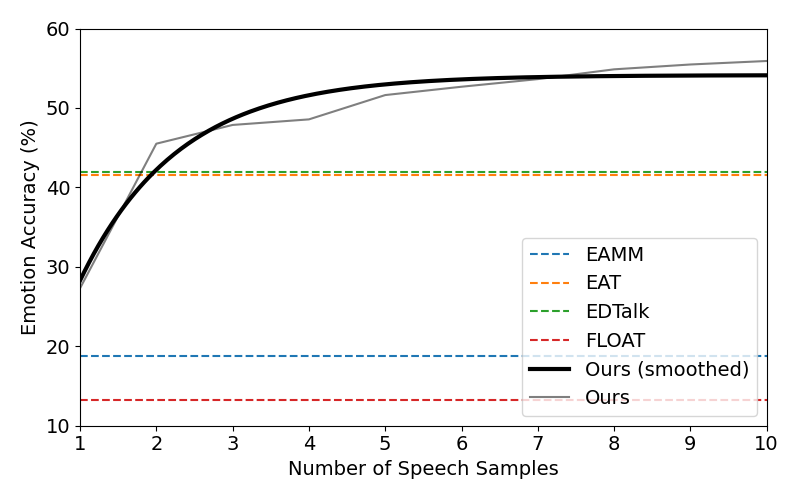}
\vspace{-5mm}
\caption{Trend of emotion accuracy as the number of emotional speech samples (“speech-shots”) increases. “Ours (smoothed)” refers to a fitted saturating exponential curve that approximates the overall trend of emotion accuracy, removing local fluctuations in raw measurements for clearer visualization..}
\vspace{-1.0em}
\label{fig:speech_shot}
\end{figure}

\noindent \textbf{Impact of speech-shot.}
Figure~\ref{fig:speech_shot} shows that our emotion accuracy steadily improves as more emotional speech samples (“speech-shots”) are aggregated, surpassing all baselines with only two samples. This improvement comes from averaging multiple speech-derived semantic vectors, which suppresses speaker-specific variations and yields a more stable emotion representation.
In contrast, all baselines remain flat because their emotion sources are fixed to ground-truth conditions—GT labels for label-based methods, GT expressive frames for image-based methods, and Speech-to-Emotion predictions aligned to GT labels for audio-based methods. These GT-driven settings already correspond to each baseline’s maximum attainable (upper-bound) accuracy, and thus additional speech samples provide no benefit. We use 10 speech-shots in all main experiments, where our performance saturates.

\begin{figure*}
    \centering
    \includegraphics[width=\linewidth]{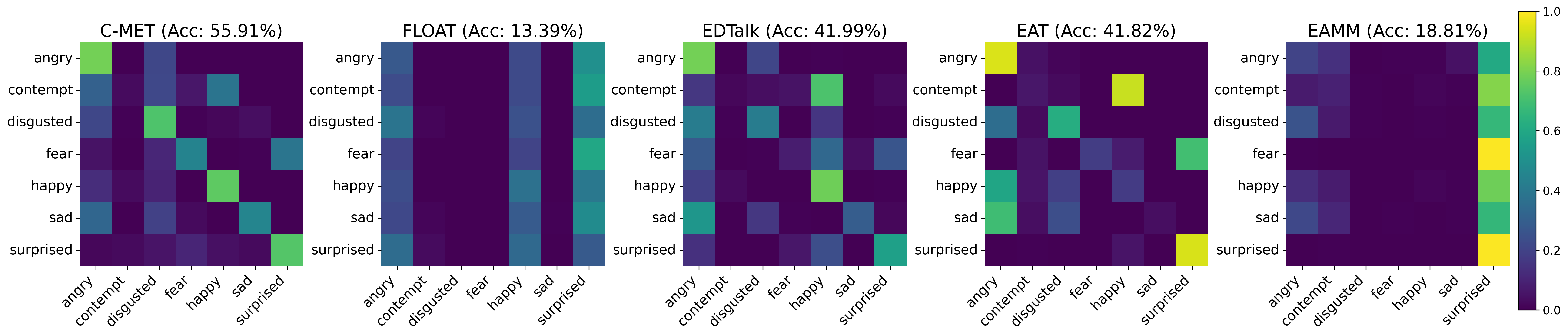}
    \caption{Confusion matrices of input (x-axis) and predicted emotions (y-axis) on MEAD across models.}
    \label{fig:confusion}
\end{figure*}

\noindent \textbf{Emotion consistency evaluation.}
Following SPACE~\cite{gururani2023space}, we evaluate emotion 
consistency using confusion matrices between input emotions and 
classifier predictions, as shown in Figure~\ref{fig:confusion}.
Our model (C-MET) exhibits the most clearly concentrated diagonal 
patterns among all compared methods, reflecting accurate and 
consistent emotion control across all seven categories.
In contrast, FLOAT shows a largely diffuse matrix 
with no discernible diagonal structure, indicating that its 
discrete speech-to-emotion module fails to reliably transfer 
the target emotion.
EAMM similarly produces scattered predictions, 
with the classifier predominantly assigning \textit{surprised} 
regardless of the intended emotion, suggesting limited expressive 
control.
EAT demonstrates a partial diagonal, yet \textit{sad} is 
frequently misclassified as \textit{angry} or \textit{disgusted}, 
suggesting that label-based generation is prone to bias toward 
visually dominant expressions and struggles to faithfully 
reproduce more subtle emotional states.
EDTalk achieves comparable accuracy to EAT but shows notable 
weaknesses in certain categories: \textit{fear} is largely 
misclassified, and \textit{happy} predictions are scattered across 
multiple emotion classes, with some confusion also observed between 
\textit{angry} and \textit{disgusted}. This is likely because 
reference image signals provide an imperfect emotion conditioning 
when editing neutral videos, as the reference may not fully capture 
the target expression. In contrast, C-MET conditions generation on 
more robust emotion semantic vectors learned in a disentangled 
space, enabling more reliable emotion transfer across diverse 
categories.

\noindent \textbf{Ablation study on audio encoder.}
Table~\ref{tab:ablation_audio_encoder} compares two audio encoder choices 
for C-MET: emotion2vec+large~\cite{ma2023emotion2vec} and 
Qwen2.5-Omni~\cite{xu2025qwen2}.
emotion2vec+large yields higher emotion accuracy---the 
primary metric of our task---as it is pretrained on large-scale 
emotion-specific speech corpora, making its representations more 
aligned with affective cues.
Furthermore, emotion2vec+large maintains lower inference latency 
compared to Qwen2.5-Omni, whose substantially larger model scale 
introduces considerable computational overhead.
Regarding the role of contrastive learning, removing 
$\mathcal{L}_\text{cnt}$ from the Qwen2.5-Omni variant leads to 
a consistent drop in emotion accuracy, which aligns with our 
finding in Table 2 of the main text that $\mathcal{L}_\text{cnt}$ 
primarily contributes to cross-modal alignment.
These results justify our choice of emotion2vec+large as the 
default audio encoder, balancing emotion accuracy and inference 
efficiency.

\begin{table}[t]
\centering
\small
\setlength{\tabcolsep}{0.5mm} 
\renewcommand{\arraystretch}{1.05}
\begin{tabular}{l|ccccc}
\toprule
 & \multicolumn{5}{c}{\textbf{Metric}} \\
 \textbf{Audio encoder} &
AITV\,$\downarrow$ &
FID\,$\downarrow$ &
FVD\,$\downarrow$ &
$\text{Sync}_{\text{conf}}\,\uparrow$ &
$\text{Acc}_{\text{emo}}\,\uparrow$ \\
\midrule
emotion2vec+large &

\textbf{2.643} & 90.804 & \textbf{329.862} & \textbf{7.9996} & \textbf{55.91} \\
Qwen2.5-Omni &
3.358 & 88.320 & 333.695 & 7.9985 & 52.06 \\
Qwen2.5-Omni w/o $L_{\text{cnt}}$ &
3.358 & \textbf{87.226} & 332.618 & 7.9592 & 51.18 \\
\bottomrule
\end{tabular}
\caption{Ablation study on audio encoder on MEAD.}
\label{tab:ablation_audio_encoder}
\end{table}

\noindent \textbf{Full metrics of ablation study.}
As shown in Table~\ref{tab:ablation_loss_full},  adding the contrastive loss \(L_{\text{cnt}}\) improves both visual quality and temporal consistency, while the direction loss \(L_{\text{dir}}\) yields the highest emotion accuracy by enforcing more discriminative emotion representations. Among all configurations, we adopt the full loss setup \(L_{\text{recon}}+L_{\text{cnt}}+L_{\text{dir}}\) for the main experiments because emotion accuracy is the most important metric for the emotion editing task, whereas visual quality metrics remain comparable across settings. This configuration therefore provides the best overall balance, maximizing emotional expressiveness without sacrificing perceptual realism.

Table~\ref{tab:ablation_network_full} summarizes the effect of integrating C-MET into disentanglement-based talking face generation models. Integrating C-MET into both PD-FGC and EDTalk consistently improves inference speed and emotion accuracy. For EDTalk, we observe slight degradations in FID, FVD, and \(\text{Sync}_{\text{conf}}\) scores; however, these differences remain comparable and were found to have negligible impact on human perception in our user study. In contrast, the gain in emotion accuracy is substantial, indicating that C-MET provides a more meaningful improvement on the primary objective of emotion editing.

Table~\ref{tab:emotion_wise_accuracy} presents the emotion-wise accuracy across seven basic emotions on MEAD and CREMA-D. Overall, C-MET (Ours) achieves the highest average accuracy on both datasets, outperforming EDTalk by a substantial margin on MEAD (55.91\% vs. 41.99\%) and maintaining the best performance even under the greater identity and recording variability of CREMA-D (43.47\%). Prior approaches often exhibit strong biases toward a limited subset of emotions—EAT, for instance, frequently produces “frowning” expressions with closed eyes, which artificially boosts accuracy for negative emotions but significantly harms performance on positive ones. In contrast, C-MET provides a more balanced treatment of the affective spectrum, achieving high accuracy for both negative and positive emotions (e.g., 78.57\% for Happy and 88.64\% for Sad). These results indicate that C-MET captures emotion-relevant dynamics in a more discriminative and semantically consistent manner, enabling robust generalization across datasets with diverse identities and affective variability.

\begin{table}[t]
\centering
\small
\setlength{\tabcolsep}{1.0mm} 
\renewcommand{\arraystretch}{1.05}
\begin{tabular}{ccc|ccccc}
\toprule
\multicolumn{3}{c|}{\textbf{Loss}} & \multicolumn{5}{c}{\textbf{Metric}} \\
 \(L_{\text{recon}}\) & \(L_{\text{cnt}}\) & \(L_{\text{dir}}\) &
AITV\,$\downarrow$ &
FID\,$\downarrow$ &
FVD\,$\downarrow$ &
$\text{Sync}_{\text{conf}}\,\uparrow$ &
$\text{Acc}_{\text{emo}}\,\uparrow$ \\
\midrule
\(\checkmark\) & & &
\textbf{2.643} & \underline{88.951} & \underline{325.926} & 7.9892 & 49.43 \\
\(\checkmark\) & \(\checkmark\) &  &
\textbf{2.643} & \textbf{88.082} & \textbf{321.961} & \textbf{8.0018} & \underline{53.46} \\
\(\checkmark\) & \(\checkmark\) & \(\checkmark\) &
\textbf{2.643} & 90.804 & 329.862 & \underline{7.9996} & \textbf{55.91} \\
\bottomrule
\end{tabular}

\caption{Quantitative results of ablation in the training loss on MEAD.}
\label{tab:ablation_loss_full}
\end{table}

\begin{table}[t]
\centering
\small
\setlength{\tabcolsep}{1.0mm} 
\renewcommand{\arraystretch}{1.05}
\begin{tabular}{c|ccccc}
\toprule
\textbf{Disentanglement} & \multicolumn{5}{c}{\textbf{Metric}} \\ 
\textbf{Network} &
AITV$\downarrow$ &
FID$\downarrow$ &
FVD$\downarrow$ &
$\text{Sync}_{\text{conf}}\uparrow$ &
$\text{Acc}_{\text{emo}}\uparrow$ \\
\midrule
PD-FGC~\cite{wang2023progressive} 
  & 1.247 & \textbf{171.464} & 937.870 & 6.7265 & 33.36 \\
\textbf{w/ Ours} 
  & \textbf{1.180} & 173.097 & \textbf{453.436} & \textbf{6.7743} & \textbf{36.82} \\
\midrule
EDTalk~\cite{tan2025edtalk} 
  & 2.827 & \textbf{76.423} & \textbf{293.904} & \textbf{8.0529} & 41.99 \\
\textbf{w/ Ours} 
  & \textbf{2.643} & 90.804 & 329.862 & 7.9996 & \textbf{55.91} \\
\bottomrule
\end{tabular}

\caption{Quantitative results of integrating C-MET into disentanglement networks on MEAD.}
\label{tab:ablation_network_full}

\end{table}

\begin{figure}[t]
\centering
\vspace{-5mm}
\includegraphics[width=0.85\linewidth]{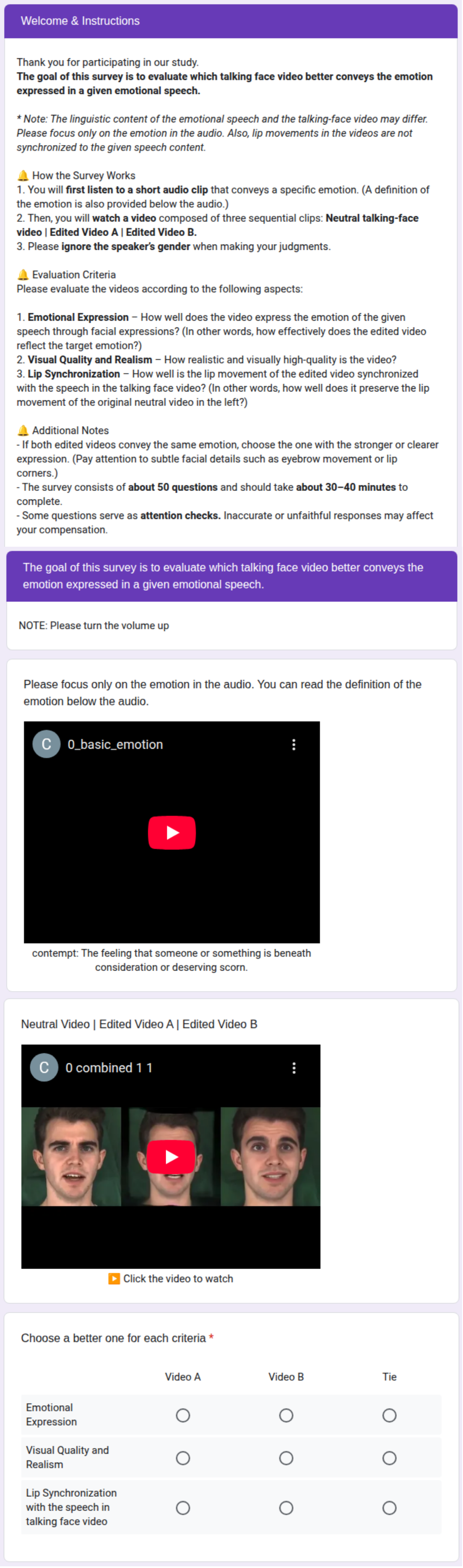}
\caption{Example interface used in our user study comparing our method with a baseline.}
\label{fig:supp_user_study}
\vspace{-5mm} 
\end{figure}

\section{Human Evaluation Template}

Our experimental setup includes a speech sample conveying a specific emotion and a video concatenating three clips (a neutral video, edited video A and edited video B). After listening to the audio, participants are asked to choose better one between three options: Edited Video A, Edited Video B, or Tie. We recruit 10 participants via Amazon Mechanical Turk (AMT) and each comparison is evaluated based on the following three criteria:

\begin{itemize}    
\item \textbf{Emotional Expression}: How well does the video express the emotion of the given speech through facial expressions? In other words, how effectively does the edited video reflect the target emotion?
\item \textbf{Visual Quality and Realism}: How realistic and visually high-quality is the video?
\item \textbf{Lip Synchronization}: How well is the lip movement of the edited video synchronized with the speech in the talking face video? In other words, how well does it preserve the lip movement of the original neutral video in the left?
\end{itemize}

Using these three criteria, we comprehensively evaluate and compare the emotional expressiveness and talking face attributes of the edited videos. To ensure diversity and fairness in evaluation, we randomly sample 50 outputs from the test set. This sampling strategy helps ensure that the results are representative and statistically meaningful.
The human evaluation template is illustrated in Figure~\ref{fig:supp_user_study}.

\section{Limitations}

Our model requires at least three pairs of neutral and emotional speech samples to achieve stable performance.
Fortunately, with recent advances in expressive text-to-speech (TTS) systems, such paired data can be easily synthesized, and existing neutral or basic-emotion speech recordings can be reused for this purpose.

Similar to other emotion-editing methods for talking-face videos, our approach does not yet handle multi-view identity images.
Consequently, its editing capability is limited when emotional modifications are needed across diverse viewpoints.
We believe that incorporating a facial expression encoder capable of multi-view reasoning could effectively address this limitation in future extensions of our framework.

In addition, current emotional talking-face datasets only support English.
As part of our future work, we plan to extend our semantic vector modeling to multilingual emotional speech data, enabling broader cross-lingual emotion generalization.

\begin{table*}[h]
\centering
\small
\setlength{\tabcolsep}{1.0mm} 
\renewcommand{\arraystretch}{1.05}
\begin{tabular}{l|cccccccc|cccccc}
\toprule
 & \multicolumn{8}{c|}{\textbf{MEAD}} & \multicolumn{6}{c}{\textbf{CREMA-D}} \\ 
\textbf{Method} & \multicolumn{8}{c|}{$\text{Acc}_{\text{emo}}\uparrow$} & \multicolumn{6}{c}{$\text{Acc}_{\text{emo}}\uparrow$} \\
 & Ang & Con & Dis & Fea & Hap & Sad & Sur & Avg & Ang & Dis & Fea & Hap & Sad & Avg \\
 \midrule
EAMM
  & 20.13 & \textbf{9.04} & 0.62 & 0.00 & 1.19 & 0.61 & \textbf{99.39} & 18.81 & 27.51 & 1.03 & \underline{41.03} & 0.32 & 23.38 & 19.15 \\
EAT 
  & \textbf{94.34} & \underline{6.02} & \underline{62.11} & \underline{17.61} & 17.26 & 3.64 & \underline{93.94} & 41.82 & \textbf{47.90} & \textbf{14.04} & \textbf{75.08} & 11.69 & 47.40 & \underline{39.97} \\
EDTalk
  & \underline{78.62} & 1.81 & 41.61 & 8.18 & \textbf{77.98} & \underline{29.09} & 56.36 & \underline{41.99} & 18.77 & 1.03 & 20.06 & \underline{30.19} & 77.60 & 29.69 \\
FLOAT 
  & 28.30 & 0.00 & 0.00 & 0.00 & 33.33 & 0.61 & 29.70 & 13.21 & \underline{45.31} & 3.77 & 13.68 & 3.57 & \underline{78.90} & 29.11 \\
C-MET (Ours) 
  & \underline{78.62} & 3.01 & \textbf{72.05} & \textbf{44.65} & \underline{75.00} & \textbf{45.45} & 73.33 & \textbf{55.91} & 4.85 & \underline{8.56} & 35.56 & \textbf{78.57} & \textbf{88.64} & \textbf{43.47} \\
\midrule
GT 
  & 98.74 & 92.73 & 56.13 & 59.35 & 100.00 & 80.61 & 83.02 & 81.88 & 89.23 & 92.98 & 82.61 & 100.0 & 72.41 & 87.74 \\
\bottomrule
\end{tabular}

\caption{Emotion-wise accuracy comparison on MEAD and CREMA-D.}
\vspace{-5mm} 
\label{tab:emotion_wise_accuracy}

\end{table*}

\end{document}